\documentclass[lettersize,journal]{IEEEtran}
\bstctlcite{bstctl:etal, bstctl:nodash, bstctl:simpurl}

\usepackage{amsmath}
\usepackage{adjustbox}
\usepackage{amssymb}
\usepackage{booktabs}
\usepackage{xspace}
\usepackage[table]{xcolor}
\usepackage{graphicx}
\usepackage{multirow}
\usepackage{pifont}
\usepackage{color}
\usepackage{cite}
\usepackage{bm}
\usepackage[figureposition=bottom,tableposition=top]{caption}

\definecolor{gray}{RGB}{142,142,142}
\definecolor{gray9}{gray}{.9}
\definecolor{gray95}{gray}{.95}
\definecolor{gray8}{gray}{.8}
\definecolor{gray85}{gray}{.85}
\definecolor{darkgreen}{RGB}{0, 150, 0}
\definecolor{darkred}{RGB}{200, 0, 0}

\newcommand{\ch}{{\color{darkgreen} \ding{51}}}
\newcommand{\xm}{{\color{darkred} \ding{55}}}

\DeclareMathOperator*{\argmax}{arg\,max}

\makeatletter
\DeclareRobustCommand\onedot{\futurelet\@let@token\@onedot}
\def\@onedot{\ifx\@let@token.\else.\null\fi\xspace}

\makeatother

\usepackage[colorlinks,pagebackref=false,citecolor=blue,bookmarks=false,hypertexnames=true]{hyperref}
\title{\LARGE \bf LiDAR-BEVMTN: Real-Time LiDAR Bird's-Eye View \\ Multi-Task Perception Network for Autonomous Driving}

\author{Sambit Mohapatra$^{1,3\dag}$, 
Senthil Yogamani$^{2\dag}$, 
Varun Ravi Kumar$^{1}$, \\
Stefan Milz$^{4}$,
Heinrich Gotzig$^{1}$ and
Patrick M\"ader$^{3}$\quad
$^\dag$co-first authors \\

{$^{1}$Valeo, Germany\quad
$^{2}$Valeo, Ireland\quad
$^{4}$SpeenLab.ai, Germany\quad
$^{3}$TU Ilmenau, Germany}
}

\begin{document}
\maketitle
\begin{abstract}

LiDAR is crucial for robust 3D scene perception in autonomous driving. LiDAR perception has the largest body of literature after camera perception. However, multi-task learning across tasks like detection, segmentation, and motion estimation using LiDAR remains relatively unexplored, especially on automotive-grade embedded platforms. We present a real-time multi-task convolutional neural network for LiDAR-based object detection, semantics, and motion segmentation. The unified architecture comprises a shared encoder and task-specific decoders, enabling joint representation learning. We propose a novel Semantic Weighting and Guidance (SWAG) module to transfer semantic features for improved object detection selectively. Our heterogeneous training scheme combines diverse datasets and exploits complementary cues between tasks. The work provides the first embedded implementation unifying these key perception tasks from LiDAR point clouds achieving 3ms latency on the embedded NVIDIA Xavier platform. We achieve state-of-the-art results for two tasks, semantic and motion segmentation, and close to state-of-the-art performance for 3D object detection. By maximizing hardware efficiency and leveraging multi-task synergies, our method delivers an accurate and efficient solution tailored for real-world automated driving deployment. Qualitative results can be seen at \url{https://youtu.be/H-hWRzv2lIY}.\par


\end{abstract}
\section{Introduction}

LiDAR sensors have become common in autonomous driving assistance sensor suites \cite{joseph2021autonomous}. Though expensive, they provide superior 3D point clouds needed for accurate 3D scene understanding \cite{kumar2018near, kumar2021svdistnet}. They are also more robust to conditions like low light and adverse weather \cite{uricar2019challenges, uricar2019desoiling, dhananjaya2021weather}.  Processing sparse 3D LiDAR point clouds is computationally intensive, and alternate 2D simplified representations such as range images and Birds Eye View (BEV) representations are commonly used for improving efficiency. There are only a few papers that explore this in the literature, and none of these focus on automotive-embedded platform implementations.\par

In deep learning, our primary focus is often on optimizing a specific metric to achieve a high score on a benchmark for a particular task. While we can reach desired accuracy through iterative fine-tuning and adjustments, neural networks for individual tasks often overlook the knowledge that could be gained from related tasks in a multi-task training process. A model's generalization capabilities for the original task can be improved by leveraging the training signals from related tasks. This methodology is called Multi-Task Learning (MTL).\par



Employing an efficient multi-task model offers numerous advantages over multiple single-task models, including improved performance in embedded systems, certification, validation, and testing. Moreover, deploying a single MTL model is more straightforward than managing several independent single-task learning (STL) models.\par

In the context of autonomous driving with LiDAR, MTL holds great importance, as it enables the exploitation of shared knowledge from various related tasks, leading to improved perception and decision-making in autonomous vehicles. Jointly trained tasks have a regularizing effect on each other, reducing overfitting issues. While achieving different end goals, features learned by tasks generally have a fair amount of overlap. This makes it possible to use features from one task as additional input to another. This is also something we have leveraged in this work. However, MTL has it's own set of challenges and needs careful design and training strategies to ensure best joint optimization across multiple tasks. Selection of tasks plays a crucial role in ensuring features can be shared across different tasks. These are elaborated in more detail below. \par

\textbf{MTL challenges:} Availability of training data for all the selected tasks is a key requirement. Unfortunately, this is not always easily available. Furthermore, the dataset sizes for different tasks are often quiet uneven. To truly leverage the benefits of MTL, it is very important to select tasks that are related and have a high overlap in feature space. However, selection of tasks is again dependent on availability of annotated datasets.
Optimizing multiple loss functions inherently gives rise to the problem of certain losses dominating others and hence not reaching the global minimum for the overall optimization objective. Therefore, it is crucial to weight the tasks appropriately to avoid optimization bias. 
With the end goal of deployment as an Autonomous Driving (AD) application, complexity and size of the network poses direct challenges to it's portability on embedded targets. Here again, the introduction of multiple tasks generally leads to bigger and more complex architectures.\par




Having discussed key challenges to MTL, this work presents a multi-task Convolutional Neural Network (CNN) for real-world use cases and deployment scenarios. Concretely, we design a CNN that performs object detection, semantic segmentation, and motion segmentation with LiDAR point clouds. The idea behind this choice of tasks is elaborated below.\par

\textbf{Task selection: }Object detection and semantic segmentation feed directly into the motion planning stage while motion segmentation plays key role in environmental modelling and Simultaneously Localization and Mapping (SLAM) applications. Several other tasks such as instance segmentation, odometry and scene flow can build upon these tasks. Therefore, these three tasks can be considered as basic components of the perception stack.
These tasks have high overlap in feature space. This is particularly true in our case. For example, we detect cars as key points which is binary classification task in the object detection pipeline. These key points can be considered a subset of semantic masks for cars from the semantic segmentation head. Therefore, semantic features are directly useful for detecting cars and vice versa. In addition, motion segmentation is chosen as a binary classification task in our design as motion is a important cue in autonomous driving scenes and can complement appearance based semantic tasks discussed above.
Furthermore, it is relatively easy to create datasets for these tasks. We use the KITTI \cite{geiger2013vision} dataset for object detection and SemanticKITTI \cite{behley2019semantickitti} for semantic and motion segmentation. While there is a disparity in number of training frames, both datasets are created with the same data collection platforms and hence are fairly uniform. We also train and validate on the Waymo \cite{sun2020scalability} dataset on two tasks for a more direct comparison with another multi-task method \cite{ye2022lidarmultinet}. \par

Our main contributions to this work include the following:
\begin{itemize}
    \item We demonstrate a real-time multi-scan multi-task CNN architecture for joint object detection, semantic segmentation, and motion segmentation from LiDAR point clouds. 
    \item We propose a Semantic Weighting and Guidance (SWAG) module that selectively transfers relevant semantic features to improve object detection accuracy.
    \item A heterogeneous training strategy combining diverse datasets and leveraging multi-task synergies.
    \item We propose a simple yet effective range-based point cloud densification technique to enrich object boundaries and improve long-range detection.
    \item We perform extensive ablation studies and demonstrate the real-time capabilities on an embedded platform - the NVIDIA Xavier AGX development kit.
\end{itemize}

\section{Related Work}

In this section, we review relevant literature on the three key LiDAR-based perception tasks addressed in our work - object detection, semantic segmentation, and motion segmentation. We first discuss recent approaches that tackle each task independently. We then examine prior multi-task methods that aim to solve subsets of the tasks jointly.

Our goal is to provide background on the single-task state-of-the-art, highlight limitations of independent modeling, and motivate the need for multi-task learning across all three modalities. This sets the stage for our proposed LiDAR-BEVMTN framework which unifies these complementary perception tasks in a joint representation learning paradigm.
\subsection{Object detection}
Object detection (OD) is finding objects of interest in a scene and localizing them with an oriented bounding box. It is typically done on a per-frame basis. However, using multiple motion-compensated past frames can improve detection accuracy significantly, particularly for slow-moving and static objects, as it allows more observations. Object detection is a crucial component of almost any AD stack since this is crucial for functions like collision avoidance, object tracking, and emergency braking. It also supports other AD tasks in achieving various end-user functions.\par

Architectures for object detection are broadly divided into 2-stage region proposal and refinement methods and single-stage methods. Due to the inherent architecture, 2-stage methods tend to be slower while generally more accurate.
Some early work like~\cite{zhou2018voxelnet, maturana2015voxnet} first group 3D point cloud into 3D tensors called voxels and then use 3D convolutions on the voxels. However, 3D convolutions have proved to be too slow for real-time operation. This approach was improved by \cite{yan2018second}, which uses sparse convolutions applied on voxel-wise features for fast inference. They also introduce a widely used data augmentation technique where object point clouds are sampled from the training set and introduced into different point clouds to solve the problem of relatively few objects in the dataset \cite{geiger2013vision}. However, at 50ms per frame inference time, they are still unsuitable for real-time operation.
Another 2-stage method that improved the Average Precision (AP) for object detection was \cite{shi2019pointrcnn}, which tries to remove the large number of anchor boxes used in traditional object detection by first learning point-wise features using~\cite{qi2017pointnet++} for segmenting the point cloud into foreground and background points and simultaneously generating a small number of proposals. The second stage then refines these proposals for 3D box regression. While introducing some novel concepts, the network relies on effective segmentation of foreground and background points which can become difficult at long range due to extreme sparsity and occlusion. This is also seen in the results' relatively significant AP drop for hard category objects.
More recently, \cite{shi2020pv} proposed another 2-stage voxel-point hybrid method that achieves state-of-the-art performance in terms of AP on the KITTI and Waymo datasets \cite{sun2020scalability}. They learn features on a point and voxel level using a 3D CNN to learn voxel-wise features and then reduce the feature space by a small set of keypoint features. More recently, \cite{wu2022sparse, borse2023x} proposed a multi-modal approach to fuse features from images and LiDAR point clouds that achieves high KITTI benchmark scores. However, our work is focused on LiDAR-only approaches.\par
Among single-stage approaches, methods like \cite{mohapatra2021bevdetnet, yang2018pixor} project 3D point cloud to 2D BEV representation and apply efficient 2D convolutions for fast object detection. 
A similar approach is followed by~\cite{simony2018complex}, which uses an Euler region proposal network for complex-valued orientation regression.
Our previous work \textit{BEVDetNet} \cite{mohapatra2021bevdetnet} achieves an impressive inference latency of 6ms per frame on Nvidia Xavier AGX. Nevertheless, a significant challenge such grid-based methods face is that the sparsity of the 3D point clouds leads to many empty cells in the grid, which negatively impacts performance. Another seminal work is \cite{lang2019pointpillars}, which uses a particular pillar-encoding layer to generate a pseudo-2D representation of point cloud and then uses Single Shot Detector (SSD) \cite{liu2016ssd} style heads for bounding box prediction. However, this additional pre-processing layer and many anchor boxes used in SSD-style detection heads are detrimental to the real inference frame rate.
Similarly, \cite{zheng2021se} also employ an SSD-based teacher-student architecture where two identical SSD networks are used as a teacher and student. The teacher is trained directly using the ground truth annotations. The student uses the teacher's predictions and the ground truth annotations to train using a consistency loss and an orientation-aware distance-IoU loss, respectively. However, the dependency on the teacher's prediction also makes the student vulnerable to bad-quality soft targets from the teacher network. Also, this dual network architecture doubles the amount of GPU memory and compute power needed and hence would only be suitable for real-time applications on embedded targets. This is evident in its relatively large inference time of 30ms per frame.
A more recent and different method is proposed by \cite{hoang2024tsstdet} which uses a vision transformer \cite{dosovitskiy2020image} based 3-stage network to perform object detection. The method overcomes incomplete point cloud and occlusions by capturing multiple rotational features and uses the pooled features with a transformer to extract robust detections. However, the multi-stage architecture suffers from increased computational time as is the case with multi-stage methods. This is evident in its 80ms runtime which is fairly large. Furthermore, scaling this architecture up to multiple tasks would increase computation and architectural complexity significantly, worsening the runtime. Another interesting approach is \cite{wu2023virtual} which uses point clouds and images in a fusion approach by using depth completion techniques (virtual points). It then reduces noise and computations using noise resistant convolution and stochastic voxel discard modules. While impressive in performance, the multi-stage approach and voxel discard layers affect inference speed and ability to deploy to embedded targets. Also, the use of depth completion using camera images does not scale very well for far off and occluded objects. This is evident from the relatively sharp drop in performance for hard objects. Finally, \cite{wu2023virtual} is a fusion method while we target a lidar only approach in this work. Similarly \cite{li2023logonet} is another lidar-camera fusion method which tries to address the issues of uniform global fusion on a whole scene level by using a 3D voxel feature extractor followed by a Region Proposal Network (RPN) \cite{yan2018second} to selectively fuse local features with corresponding semantic features from a 2D feature extractor for images. They also perform global feature fusion for non-empty voxels and finally combine local and global features in a self-attention stage to produce final predictions. They report impressive scores on the KITTI and Waymo datasets. However, they report a relatively high runtime of 100ms. This is again attributed to the multi-stage architecture and 3D operations in the voxel feature extractor.\par
\subsection{Semantic Segmentation}

Semantic segmentation refers to assigning a semantic label to each pixel or point in the scene. This way, each observation is labeled into one of the predetermined semantic classes. Again, this can be done on single frames or using several past frames. This is useful for automotive applications such as free space estimation, drivable area estimation, path planning, lane marking, and general refinement of object detection \cite{dutta2022vit, rashed2019motion}.
As with object detection, we mainly focus on LiDAR-only methods and include published work using the SemanticKITTI \cite{behley2019semantickitti} dataset. \cite{wu2018squeezeseg, wu2019squeezesegv2} transformed 3D point cloud into 2D representation using spherical projection and then used a\cite{iandola2016squeezenet} based CNN followed by refinement using conditional random field \cite{zheng2015conditional} module for semantic segmentation. This was improved upon by\cite{milioto2019rangenet++}, which uses a similar spherical projection called range images and a\cite{redmon2018yolov3} based architecture to perform fast semantic segmentation. However, range images (spherical projection used in all these approaches) do not allow easy fusion of features with that of the grid-based 2D representations typically used for object detection networks in a multi-task approach. Another novel approach was proposed by~\cite{hou2022point} where a teacher-student architecture is employed. The much larger teacher model is used to collect point and voxel-wise features from the input point cloud, and this information is then distilled into the smaller student model. Concretely, they divide the point cloud into a set of voxels and then take a subset based on distance and difficulty criteria. They then compute a similarity loss between the point and voxel features. However, the use of 3D convolutions is not suitable for real-time operation and does not fit the case of multi-task learning with 2D grid-based inputs. Furthermore, the teacher-student architecture makes joint training difficult, as in our case. 
{\cite{kong2023lasermix} presented a newer and different approach which leverages relation between traffic objects and their spatial location in a laser scan to predict semantic labels with fewer training labels in a semi-supervised way. However, their teacher-student architecture and 3D processing in the initial stages makes the network computationally expensive and cumbersome to train. Furthermore, the spatial priors have to evolve in rapidly changing traffic scenarios which will degrade performance. \cite{zhang2023growsp} presented a novel approach for unsupervised semantic segmentation using per-point 3D feature computation and successive grouping into super-points followed by semantic clustering. While it shows impressive performance in indoor settings, it is far from other supervised methods in outdoor scenarios as observed in their benchmark. More recently, \cite{li2024tfnet} improved semantic segmentation in range images by addressing the problem where several points in a point cloud map to the same location in the range image due to the limited resolution of range images. They leverage the temporal aspect of point due to motion and employ a max voting scheme to refine the wrong labels. However, this voting scheme can produce wrong results in situations where there is incremental motion across successive frames. Furthermore, representations such as BEV address the limited resolution problem in a better way.}\par
\subsection{Motion Segmentation}

Motion segmentation is a particular case of semantic segmentation where there are only two classes - motion and static, denoting pixels or points that are dynamic or static, respectively. However, in contrast to semantic segmentation, motion segmentation needs at least two successive frames since motion cannot be inferred from a single observation. Typically, many consecutive frames are used to achieve higher accuracy in motion segmentation. Motion segmentation finds key uses in systems like SLAM, object tracking, collision avoidance, and path planning systems of the AD stack.
While we have discussed several approaches for semantic segmentation, there needs to be more work for moving object segmentation - a more specialized case of semantic segmentation. One of the most notable is \cite{chen2021moving}, which uses multiple frames of point clouds converted into a spherical coordinate system (range image) to segment moving pixels. The authors motion compensate the frame and compute successive frames' differences, which helps provide motion cues. A standard encoder-decoder architecture is then used to predict a motion mask for the most recent frame. While simple and effective, their best results are achieved with eight past frames, again prohibitive from the memory and computing perspective of embedded targets in real-world applications. Also, the range view skews the aspect ratio, making fusing learned features across tasks in a multi-task scenario very hard. We use a similar idea in our previous work~\cite{mohapatra2021limoseg}, using only two past frames in a BEV space to segment motion pixels. More recently, \cite{sun2022efficient} proposed a dual encoder approach to extract spatial and temporal features and fuse them using a sparse 3D convolution module. They show improved accuracy, but the dual encoder architecture directly opposes a multi-task approach, where having a single encoder is one of the critical requirements. Also, the use of 3D convolutions is not suitable for low latency, and this shows in their rather large run times. {\cite{zeng2024mambamos} proposed a novel approach which uses multiple frames along with their timestamps and explicitly combining temporal and spatial features in a motion aware model that weights temporal features. However, for optimal temporal features, the number of past frames has to be relatively large which directly translates to larger memory and compute requirements.} \par
\subsection{Multi-task LiDAR perception}

Compared to other modalities such as image-based~\cite{leang2020dynamic, klingner2022detecting}    and fusion-based approaches \cite{liu2022bevfusion} for multi-task perception networks, there are rather few methods for LiDAR only multi-task perception. Our choice of LiDAR-only multi-task perception is based on the fact that LiDAR as a sensor sits in a sweet spot between sensors such as cameras which produce dense data but are severely affected by environmental conditions such as ambient light, rain, fog, etc., and radars which have the longest range but produce data that is too sparse for tasks like object detection and semantic segmentation. Furthermore, we tailor our application with real-time deployment with embedded hardware in mind and choose a 2D BEV representation. This enables us to use fast 2D convolutions in the architecture and also makes the output readily usable with other AD functions, such as path planning, with very little post-processing needs.
In fact, from our experiments, BEV is even better than range-image (spherical coordinates) representation at the post processing side. Inference results are directly available in ego coordinates and recovering 3D data is as simple as reading the desired pixels from BEV directly. Furthermore, most state of the art optimizing compilers such as Nvidia Tensorrt, Microsoft Onnxruntime, Apache TVM etc., provide best support and performance for the 1D and 2D operations available in frameworks like Pytorch and Tensorflow. While BEV also suffers from sparsity, for tasks like motion segmentation which need multiple past frames, BEV offers a very compact representation without additional memory and compute requirements. Most other LiDAR methods use 3D operations such as voxels and sparse 3D convolutions which are either not natively supported or highly inefficient for deployment in automotive embedded systems.
Among multi-task LiDAR-only approaches, \cite{ye2022lidarmultinet} uses a single point cloud and performs object detection, semantic segmentation, and panoptic segmentation. They employ 3D sparse convolutions as feature extractors and then project features to a 2D BEV space for contextual feature aggregation. This is then fed to task heads in BEV space for final outputs. Another interesting approach is \cite{duffhauss2020pillarflownet}, which does object detection and scene flow using two consecutive frames of LiDAR point clouds. However, they follow a two-encoder architecture for feature extraction from each stream and then fuse them using a joint encoder stage. This is inefficient from a computation latency perspective. Similarly, \cite{feng2021simple} uses an encoder-decoder architecture with sparse 3D convolutions for object detection and segmentation of roads into different parts. {Recently \cite{chen2024joint} proposed a method for scene flow and motion segmentation without using vehicle odometry directly. They reason that motion segmentation and scene flow are related tasks and have significant feature space overlap. Features are computed over 2 sequential frames and then a cosine distance function is used to estimate an initial flow mask. This is then refined using motion labels from previous scan to get the final flow vectors. However, as the number of past frames increases, 3D feature computation and point wise matching becomes computationally expensive.}\par
\section{Proposed Method} 
\label{sec:network-details}
\subsection{System Architecture}

\begin{figure}
    \centering
    \includegraphics[width=\columnwidth]{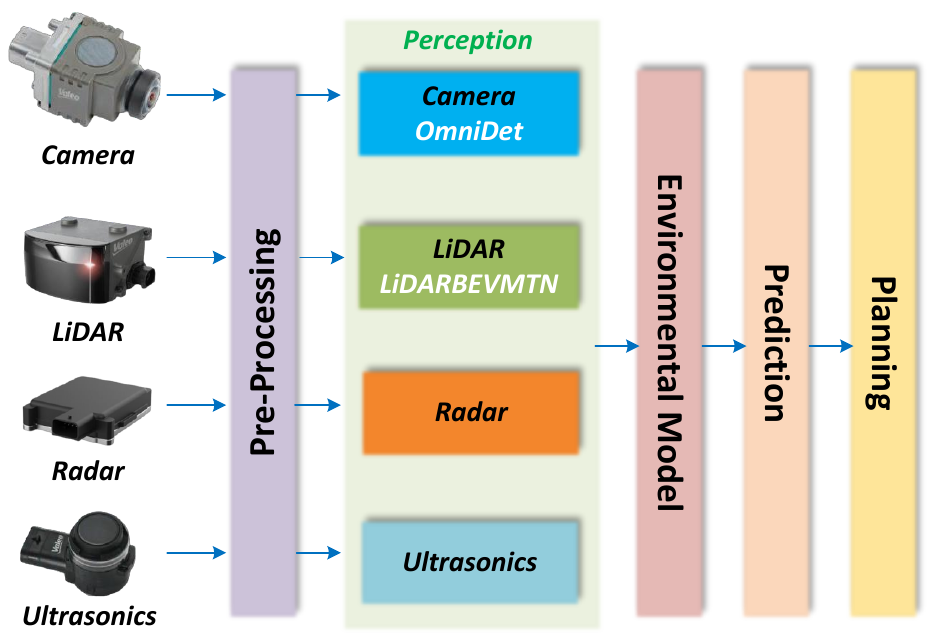}
    \caption{\bf High-level Architecture of Level 2/Level 3 Autonomous Driving System.}
    \label{fig:sys-arch-ad}
\end{figure}
Fig.~\ref{fig:sys-arch-ad} shows a typical Level 3 AD system architecture. It shows the most common combination of ultrasonic, camera, lidar and radar sensors for short to ultra-long range sensing. After pre-processing such as in \cite{elamir2022deep}, sensor data passes into the perception stage, where individual modalities pass through their perception algorithms and, optionally, fusion-based algorithms. Our work (LiDAR-BEVMTN - shown in figure) builds on our previous work \textit{OmniDet} \cite{kumar2021omnidet}, which is a multi-task camera perception model. Perception outputs feed the environmental model, which constructs a detailed model of the ego vehicle state and other actors in the environment. Tracking and motion planning algorithms then predict the next states and provide motion planning. Having established the overall picture, the following sections explain our work in greater detail.\par

\subsection{Input Pre-processing}

The BEV representation transforms 3D point clouds into a 2D grid structure while preserving key information. Each point $\mathbf{p} = (x, y, z)$ is mapped to a cell $(u, v)$ in the BEV image $I_{BEV} \in \mathbb{R}^{H\times W}$ based on its x-y location:
\begin{align}
u = \left\lfloor\frac{x - x_{min}}{r_x}\right\rfloor & v = \left\lfloor\frac{y - y_{min}}{r_y}\right\rfloor
\end{align}
where $(x_{min}, y_{min})$ are the lower bounds of the BEV range, and $(r_x, r_y)$ define the cell resolution. Point features like depth, intensity, etc., are aggregated in the corresponding cell. This enables efficient 2D convolutions without losing substantial information.\par

Crucially, BEV preserves aspect ratios as objects maintain consistent dimensions with distance. The compact BEV representation also simplifies integration with downstream modules for planning and prediction. As in \cite{mohapatra2021bevdetnet}, we use a $60m \times 60m$ BEV range with $\mathbf{0.125m}$ resolution, filtering points outside the view. The resulting $I_{BEV} \in \mathbb{R}^{480\times 480\times C}$ retains high-fidelity scene representations for multi-task learning.\par
\begin{figure}[t]
\captionsetup{singlelinecheck=false, font=small, skip=0pt, belowskip=-10pt}
    \centering
    \includegraphics[width=0.50\textwidth, height=0.25\textwidth]{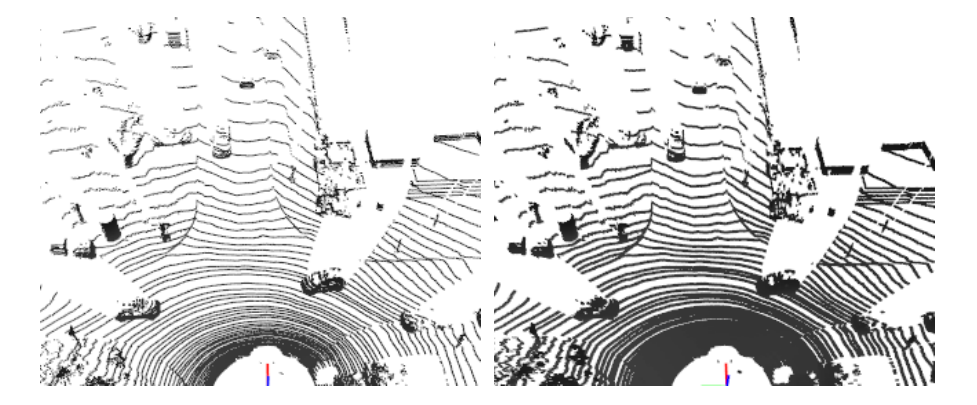}
    \caption{\textbf{Illustration of proposed Range based point cloud densification} - Raw point cloud (left) and densified point cloud (right).}
    \label{fig:osmnet_sparse_bev}
\end{figure}
\begin{figure}[t]
\captionsetup{singlelinecheck=false, font=small, skip=0pt, belowskip=-12pt}
    \centering
    \includegraphics[width=0.49\textwidth, height=0.45\textwidth]{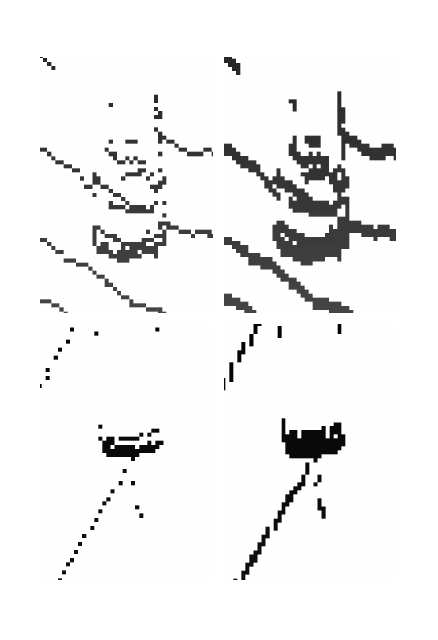}
    \caption{\textbf{Effect of point densification on objects in BEV} - BEV raw (left) and BEV densified (right).}
    \label{fig:point_aug_bev}
\end{figure}

\textbf{Range-based point cloud densification:} We propose a simple yet effective range-based point cloud densification technique that can improve point density along the depth dimension even at test time. We convert every 3D point $\mathbf{p} = (x, y, z)$ to its spherical coordinates - range ($r$), azimuth ($\phi$), and elevation ($\theta$):
\begin{alignat}{2}
r &= \sqrt{x^2 + y^2 + z^2} \label{eq:r} \\
\theta &= \sin^{-1}\left(\frac{z}{r}\right) \quad &&\text{elevation angle} \label{eq:theta} \\
\phi &= \tan^{-1}\left(\frac{y}{x}\right) \quad &&\text{azimuth angle} \label{eq:phi}
\end{alignat}
We then increase range by $\Delta r \in [0.1, 0.3]$ and recompute the new densified points' Cartesian coordinates $(x', y', z')$ as:
\begin{align}
x' &= (r + \Delta r)\cos{\theta}\sin{\phi} \label{eq:xprime} \\
y' &= (r + \Delta r)\cos{\theta}\cos{\phi} \label{eq:yprime}\\
z' &= (r + \Delta r)\sin{\theta} \label{eq:zprime}
\end{align}
Since only the range is modified, the new points trace along the depth of objects, improving density in the depth dimension. 
For small increment in range, the object boundaries and contours are not disturbed since new points land inside object body. The azimuth and elevation angles for original points are maintained throughout.
This is particularly beneficial when projected to BEV, enhancing visibility along the depth axis. However, over-augmentation can cause points to leave object boundaries. This is particularly true for objects like pedestrians and bicyclists with small volume and few points. Another potential problem with over-augmentation is ground points can leave x-y plane and move below it. Also, it does not truly increase resolution or add new surface details like true upsampling methods. Furthermore, performance may degrade if the test data contains less noise than the training data, as the model would overfit to the artificial noise distribution. We conduct ablation experiments for object detection, as this augmentation modifies ground truth for segmentation. Fig.~\ref{fig:osmnet_sparse_bev} illustrates the point densification effects. Fig.~\ref{fig:point_aug_bev} demonstrates the effect in BEV. The two objects are clearly densified while maintaining contours and boundaries. This method can be further refined to avoid densifying ground points using point normals. However, for object detection, this step is not necessary. The proposed technique provides a simple but effective way to enrich point density at test time.\par

\subsection{Network Architecture}

\begin{figure*}
\captionsetup{singlelinecheck=false, font=small, skip=2pt, belowskip=-8pt}
        \centering
        \includegraphics[width=\textwidth]{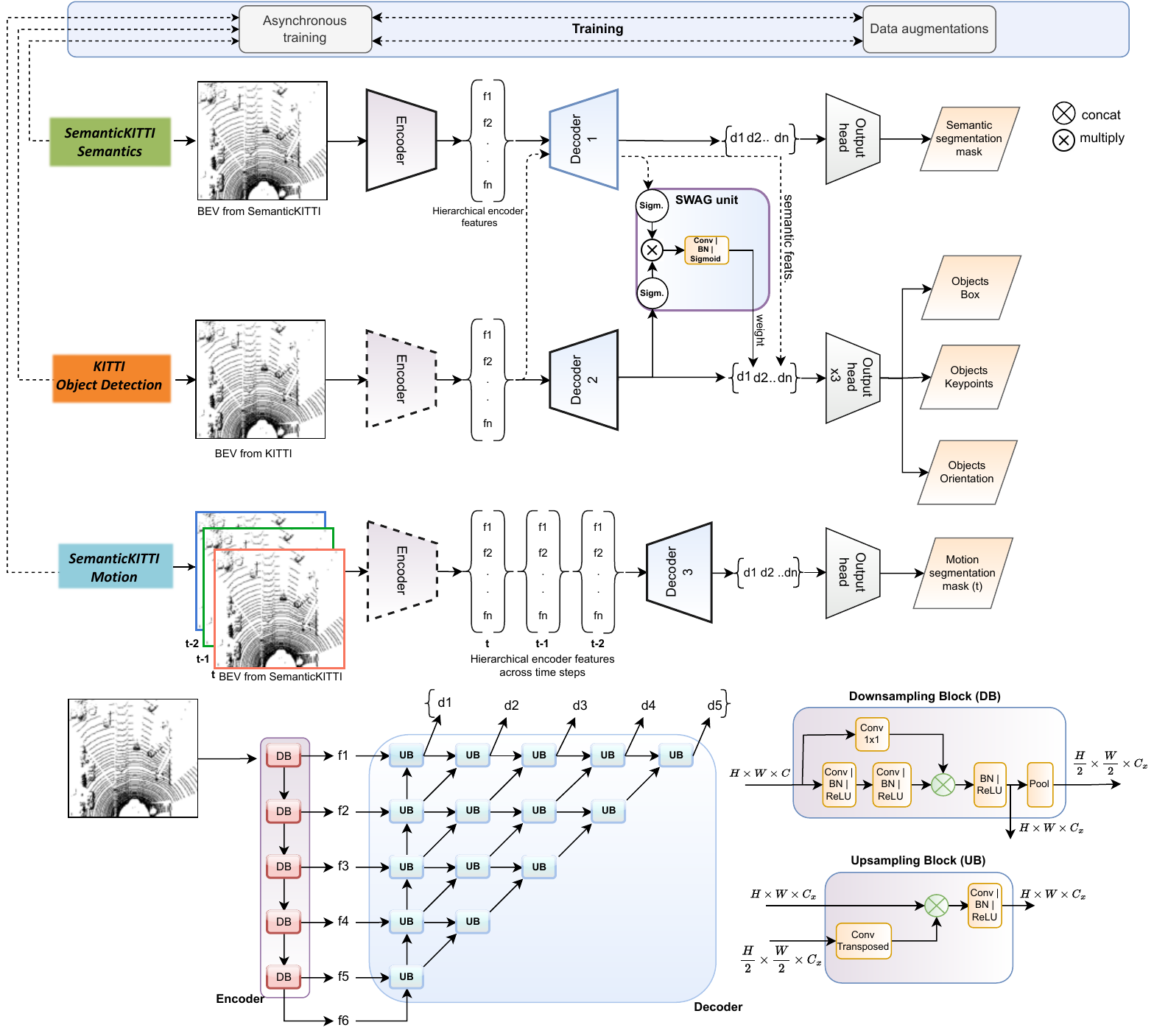}
        \caption{\textbf{
        LiDAR-BEVMTN multi-task learning architecture.} Top: High-level view showing shared encoder and task-specific decoders. Bottom: Encoder and decoder sub-modules details. Key features: 1) Shared encoder for unified feature extraction 2) SWAG module for cross-task interactions 3) Asynchronous training across datasets 4) Task-specific decoders enable multi-task pixel-level prediction.}
        \label{fig:Figure_overall_nw_bd}  
\end{figure*}

Our network adopts an encoder-decoder CNN backbone for feature extraction, followed by task-specific decoders and output heads (Fig.~\ref{fig:Figure_overall_nw_bd}). The encoder comprises a shared multi-scale feature extractor based on a ConvNeXt-style architecture. It outputs a pyramidal hierarchy of features that encode increasing semantic complexity while reducing spatial dimensions.\par

The core design principle is to leverage complementary cues from the different tasks during the encoder and decoder stages. The encoder features are directly fed to a decoder and pixel-wise classification head for semantic segmentation.\par

For object detection, we introduce a SWAG unit that selectively incorporates semantic features to improve detection performance. Specifically, the encoder features are fed to the semantic and detection decoders. The SWAG module computes a relevance match between intermediate features from both branches. This match vector is used to weight the semantic features before fusing them with the detection features. By learning to select useful semantic cues, SWAG provides improved generalization. Our box regression scheme based on key points representing object centers and orientation and dimension prediction heads allows high-density prediction while preventing collisions.\par

As in our previous work~\cite{mohapatra2021bevdetnet}, we represent objects as key points, where a key point $\mathbf{p}_i \in \mathbb{R}^2$ marks the $i^{th}$ object's center location in the BEV image. This allows high-density prediction while preventing collisions, as the object's body ensures overlap does not occur.

Corresponding to each key point $\mathbf{p}_i$, we also predict the object's orientation $\theta_i$ as a binned classification between $[0, 180]$ degrees with a bin size of $\Delta \theta = 5$ degrees:
\begin{equation}
\theta_i = \argmax_j \{\mathbf{z}_{ij}\} \quad \text{where} \quad j \in \{0, \frac{180}{\Delta \theta} - 1\}
\end{equation}
where $\mathbf{z}_i \in \mathbb{R}^{180 / \Delta \theta}$ is the predicted classification logit vector for object $i$.
Additionally, a box dimension prediction head is added to regress the width $w_i$ and length $l_i$ for each detected object $i$:
\begin{equation}
w_i = f_w(\mathbf{v}_i) \quad l_i = f_l(\mathbf{v}_i)
\end{equation}
where $\mathbf{v}_i$ contains box regression features and $f_w, f_l$ denotes width and length prediction layers. As the semantic segmentation head trains, its accuracy improves over time. This enhances the quality of semantic features contributed to the object detection head via the proposed SWAG unit, leading to increased detection performance.\par

Motion segmentation operates on multi-frame input, with the encoder applied individually to each frame. The resulting features are concatenated temporally and compressed before feeding to the motion decoder and segmentation head. Unlike semantics, motion patterns can be negatively impacted by semantic errors; hence semantic guidance is excluded.\par

The proposed architecture unifies multiple perceptual tasks through shared encoding, selectively leverages cross-task interactions, and employs task-specific decoding to balance joint representation learning with specialized prediction. This achieves significant performance gains in each task compared to independent modeling.\par
\textbf{Encoder and Decoder:} The encoder comprises 5 cascaded downsampling blocks (DB) as seen in Figure \ref{fig:Figure_overall_nw_bd} (bottom right) based on a modified ConvNeXt-style \cite{liu2022convnet} architecture. Each DB contains convolutional layers followed by batch normalization and ReLU activation, configured in a residual block structure. Compared to a standard ConvNeXt block, the key modification is that two outputs are generated from each DB - one retaining the original spatial resolution and another with 2x downsampling applied.\par

Specifically, each DB first applies a strided 3x3 convolution with stride 2 to downsample the input feature map. This downsampled output is then fed to the next DB in the cascade. Meanwhile, the original high-resolution input goes through a residual path consisting of two 3x3 convolutions with batch normalization and ReLU. This produces an output feature map with the same resolution as the input. After the cascaded DB stages, the encoder outputs a set of multi-scale feature maps consisting of the high-resolution outputs from each DB and the low-resolution output of the final DB. This pyramidal hierarchy encodes rich semantic features while preserving spatial details.\par

The decoder module adopts an expansive pathway that selectively progressively fuses the encoder feature maps. It consists of a series of upsampling and concatenation blocks. Each block upsamples the lower-resolution input using a transposed convolution layer to match the size of the corresponding high-resolution encoder output. The upsampled features are then concatenated channel-wise with the encoder output before going through additional convolutional layers. This learnable feature fusion allows the combining of semantics from the downsampled stream with spatial details from the high-resolution stream. The progressive fusion through the decoder recovers fine-grained spatial information while leveraging context from the encoder. The final output is a dense feature map with the same spatial dimensions as the input image, which can be tasked with pixel-level prediction.\par

Overall, the proposed encoder-decoder architecture provides an effective backbone for multi-task dense prediction, achieving a balance between semantic abstraction and spatial precision.\par

\textbf{Semantic Weighting and Guidance (SWAG) unit:} The goal of the SWAG unit is to introduce a weighted set of semantic features produced by feeding object detection encoder (the encoder is still the same for all tasks) features into the semantic decoder. A simple but naive approach to introducing semantic features into object detection would be just concatenating them together. However, not all features from the semantic head are equally helpful for the object detection head. Let $f_{\text{sem}} \in \mathbb{R}^{H\times W \times C}$ and $f_{\text{od}} \in \mathbb{R}^{N\times D}$ denote the intermediate feature maps from the semantic segmentation and object detection decoder branches respectively, where H, W are spatial dimensions, C is number of semantic channels, N is number of candidate boxes, and D is the feature dimensionality for each box. To compute the relevance match, we first apply MLPs to project each into a joint K-dimensional embedding space:
\begin{align}
h_{\text{sem}}(f_{\text{sem}}) &= \sigma(W_{1}^{\text{sem}} f_{\text{sem}} + b_{1}^{\text{sem}}) \\
h_{\text{od}}(f_{\text{od}}) &= \sigma(W_{1}^{\text{od}} f_{\text{od}} + b_{1}^{\text{od}})
\end{align}
Where:
$h_{\text{sem}}()$ and $h_{\text{od}}()$ are the MLP projections
$\sigma$ is the sigmoid activation function
$W_{1}^{\text{sem}}$ and $W_{1}^{\text{od}}$ are the projection weight matrices
$b_{1}^{\text{sem}}$ and $b_{1}^{\text{od}}$ are the bias terms
$f_{\text{sem}}$ and $f_{\text{od}}$ are the input feature maps.

Our proposed SWAG module computes a match $m$ between object detection features and semantic output:
\begin{equation}
m = \sigma(h_{od}(f_{od})) \odot \sigma(h_{sem}(f_{sem}))
\end{equation}
where $h_{od}$ and $h_{sem}$ are small MLPs, $\sigma$ is the sigmoid function, and $\odot$ is element-wise multiplication between the embeddings. MLPs are induced to provide learnable nonlinear projections that map detection and segmentation features to a common embedding space where their joint relationships can be effectively modeled and leveraged to improve detection performance. The learnable nature and joint trainability are key advantages over using a fixed projection. This match vector captures the relevance of each semantic feature to object detection.  To convert the match to a set of learnable weight vectors $w$, we apply another conv layer:
\begin{equation}
w = \sigma(CB(m))
\end{equation}
where $\text{CB}(\cdot)$ denotes a convolution + batch norm layer, and $w \in \mathbb{R}^{C}$ contains the weighting coefficients. The weighted semantic features $f'{sem} = w \odot f{sem}$ are then 
concatenated with $f_{od}$ for the object detection head, as shown in Fig. \ref{fig:Figure_overall_nw_bd}.  The learnable projections and weightings allow SWAG to focus on useful semantic cues automatically. In this way, SWAG learns to select the most relevant semantic features to incorporate into object detection, improving generalization. While SWAG is not explicitly used on the other branches, it generally affects car segmentation across all tasks due to the shared encoder.\par

\subsection{Training Strategy}
\label{sec:input-process}

A key contribution of our work is a heterogeneous training scheme that unifies diverse datasets across multiple perception tasks. We leverage KITTI~\cite{geiger2013vision} for object detection and SemanticKITTI \cite{behley2019semantickitti} for semantic and motion segmentation. To handle differences in dataset scale and image sequence, we propose a tailored strategy:

\begin{itemize}
\item For object detection, each KITTI frame is converted to BEV and used to train the detection branch.
\item Semantic segmentation is trained on every $5\textsuperscript{th}$ SemanticKITTI frame in BEV view. This balances the sample count across tasks.
\item For motion segmentation, we identify SemanticKITTI frames with $>\delta$ moving pixels. Each frame and its past two motion-compensated frames are used as inputs, as described in~\cite{mohapatra2021limoseg}.
\end{itemize}

A key novelty is computing pseudo-residuals to enable joint encoding. For motion frames, residuals between the current and past frames highlight dynamics. For static frames, we self-multiply the depth channel and normalize it to create disparity, improving detection and semantics. The benefit of doing this rather than just repeating the depth channel is that multiplying increases the disparity between high and low height points due to decimal multiplication and helps detection and segmentation.\par

The proposed heterogeneous training uniquely combines diverse data sources and leverages multi-task interactions to enable unified perception. Our scheme balances sample counts and creates cross-task synergies within a multi-sensor setting.\par
\textbf{Asynchronous Training}: We propose an asynchronous training strategy since SemanticKITTI and KITTI have unequal training frames for object detection, semantic segmentation, and motion segmentation. For object detection, we use $\mathbf{5,957}$ labeled training frames and apply sampling data augmentations to get a total of $\mathbf{9,400}$ training frames. $\mathbf{9,560}$ training frames are used for semantic segmentation, and for motion segmentation, we have $\mathbf{4,719}$ training frames. To handle this imbalance, we adopt an asynchronous training approach. The batch index in our unified data loader runs over the most extended dataset - $\mathbf{9,400}$ frames for object detection. We remap the index for semantic segmentation by taking a modulo with $\mathbf{9,560}$; for motion segmentation, we take a modulo of the batch index with $\mathbf{4,719}$. This way, the object detector sees the entire dataset in each training epoch while the semantic and motion networks cycle through their smaller labeled sets multiple times. The three models are trained jointly in an end-to-end manner despite the unequal supervision. We find this asynchronous strategy crucial to train the multi-task model effectively. By aligning the batch indices asynchronously, we can maximize the utilization of the available supervision for each task. Our experiments validate the benefits of this tailored asynchronous training approach for joint perception on KITTI and SemanticKITTI. The index remapping is expressed as:
\begin{equation}
index_{remapped} = index \pmod{L_{dset}}
\end{equation}
Where:
\begin{list}{$\bullet$}{\leftmargin=2pt}
\item $index \in [0, N]$: Batch counter index, ranging from 0 to N
\item $N$: Length of largest dataset
\item $index_{remapped}$: Remapped index for current dataset
\item $L_{dset}$: Length of current dataset
\end{list}

We also perform data augmentation, which is crucial for improving the generalization and robustness of deep learning models. However, the effectiveness of augmentation techniques can vary across different tasks. To address this, we adopt a \emph{progressive data augmentation} strategy that gradually increases the complexity of augmentations as the training progresses. Concretely, we divide our data augmentations into task specific complex augmentations that are tailored towards individual tasks and simpler general augmentations, targeting joint tasks. Under task specific augmentations, we apply 3D point augmentation, and sampling augmentation for object detection. For segmentation tasks, we apply left-right flip, local object rotations and static object translation along x-axis (motion segmentation only) for segmentation tasks.  
For general augmentations we apply simpler methods like global rotations, scaling and random point drop (30\%).
Initially we only apply the simpler general augmentations which are applicable to all tasks and help the model learn the most number of shared features across all tasks. As training progresses, we introduce the more extreme tasks specific augmentations which effectively refine the features learnt by the later stages of decoders and output heads to make individual task heads better. Furthermore, due to the large difference between training frames, task specific augmentations are not applied to every frame of individual tasks. Table~\ref{tab:data-aug-static-progressive} demonstrates the effect of this progressive augmentation strategy. As can be seen, while most of the tasks benefit from this, motion segmentation gets the maximum improvement in performance while using progressive over static augmentations. A key reason for this is motion segmentation has the least number of training frames and our strategy of initial general augmentations help the network identify cars and other likely-mobile objects which is followed by late stage augmentation which helps the network learn the features for mobile objects from their multiple instances.

\section{Experimental Results}

\subsection{Implementation}

We employ Pytorch~\cite{paszke2017automatic} and follow a joint multi-task learning strategy to train the object detection, semantic segmentation, and motion segmentation tasks concurrently. A total of $\mathbf{3}$ BEV frames is input to the network - one for each task from their respective datasets. Each BEV has the same set of feature channels. The BEV consists of $\mathbf{3}$ successive frames for motion segmentation and has three times the number of channels. Internally, the network isolates these and computes features for each time step, as discussed in the previous section.\par

Asynchronous batch sampling accounts for the large disparity between KITTI and SemanticKITTI labeled frames. We subsample $\mathbf{1524}$ KITTI validation frames for object detection and use sequence 08 from SemanticKITTI for segmentation. For generalizability, we report semantic segmentation on the most common classes in KITTI. Also, since cars make up the majority of object types in the KITTI dataset, we train for car detection on it. This also affects the other tasks due to the shared architecture. Hence, we report results for the most commonly occurring classes out of the $\mathbf{19}$ classes generally seen in the semantic segmentation benchmark on SemanticKITTI. The training uses the Adam \cite{kingma2014adam} optimizer with a cosine annealing learning \cite{loshchilov2016sgdr} rate over $\mathbf{120}$ epochs on an Nvidia A100 GPU.\par
\subsubsection{Losses}

We use Focal loss \cite{lin2017focal} for all classification tasks, $L_{cls}$ - object key point head $(L_{kp})$, object orientation head $(L_{rot})$, semantic segmentation head $(L_{sem})$ and motion segmentation head $(L_{mot})$ and, smooth-L1 loss \cite{girshick2015fast} is used for all regression, $L_{regr}$ - object box head $(L_{box})$. The regression loss is only computed for key point locations by masking off other locations in both the ground truth and the predictions.
The total loss is then computed as a weighted sum of all losses. We adopt a rotating weighting strategy where the weights are selected from a fixed set of weights proportional to the magnitude of each loss. After every n-epochs, ratio of each task specific loss is computed with respect to the total loss and based on the magnitudes, tasks with higher losses are assigned larger weights and vice versa. This ensures that the overall joint minimization of losses progresses in a balanced manner. In our experiments, we start with more frequent weight updates in the early stages of training and weights are cycled less frequently as the network converges. We drew inspiration for this strategy from the cosine annealing learning rate which adopts a similar strategy to avoid local minima and explore entirety of the weight space.\par
\begin{equation} \label{loss_cls}
L_{cls} = -(1 - p)^{\gamma}\log(p)
\end{equation}
Where $L_{cls}$ is the classification loss, $p$ is the probability of correct classification, and $\gamma$ is a hyperparameter.
\begin{equation}\label{loss_regr}
    L_{regr} = 
    \begin{cases}
    0.5(Y_{hat} - Y)^2 / \beta, & \text{$ |Y_{hat} - Y| < \beta $}\\
    |Y_{hat} - Y| - 0.5\beta, & \text{otherwise}
    \end{cases}
\end{equation}
Where $L_{regr}$ is the regression loss, $Y_{hat}$ is the predicted value, $Y$ is the target value, and $\beta$ is a hyperparameter.
\begin{equation} \label{loss_total}
L_{total} = \sum_{i=1}^{5} w_{i} L_{i}
\end{equation}
Where $L_{i} \in {L_{kp}, L_{box}, L_{rot}, L_{sem}, L_{mot}}$ are the task-specific losses, and $w_{i} \in [0.98, 0.95, 0.90, 0.85, 0.80]$ are the loss weights.
\subsection{Evaluation}

Our experiments compare STL against MTL for object detection, semantic segmentation, and motion segmentation on the KITTI dataset in Table~\ref{tab:table1-mtl-stl}. For STL, each task is trained independently. With MTL, the tasks are trained jointly with shared parameters in the encoder backbone.\par

\begin{table}[ht]
\captionsetup{singlelinecheck=false, font=small, skip=2pt, belowskip=-2pt}
\centering
\caption{\textbf{Quantitative comparison of STL and MTL} - Object detection (Average Precision (AP)), Semantic Segmentation (IoU), Motion segmentation (IoU) and Inference speed in Frames Per Second (FPS). Object detection
and Motion segmentation show the largest gains from MTL.}
\label{tab:table1-mtl-stl}
\centering
\begin{adjustbox}{width=\columnwidth}
\setlength{\tabcolsep}{0.2em}
\begin{tabular}{@{}ccccccccccc@{}}
\toprule
\multicolumn{3}{c|}{Tasks} & 
\multicolumn{3}{c|}{Object Detection} & 
\multicolumn{4}{c|}{Semantic Segmentation} & 
\multicolumn{1}{c}{} \\ 
\cmidrule(lr){1-10}
\textit{OD} & 
\textit{SS} & 
\multicolumn{1}{c|}{\textit{MS}} & 
\cellcolor[HTML]{a5eb8d}\textit{Easy} & 
\cellcolor[HTML]{96bbce}\textit{Moderate} & 
\multicolumn{1}{c|}{\cellcolor[HTML]{fc4538}\textit{Hard}} & 
\cellcolor[HTML]{00b0f0}\textit{Car} & 
\cellcolor[HTML]{e5b9b5}\textit{Road} & 
\cellcolor[HTML]{a5a5a5}\textit{Building} & 
\multicolumn{1}{c|}{\cellcolor[HTML]{00b050}\textit{Veg.}} &
\multicolumn{1}{c}{\multirow{-2}{*}{\begin{tabular}[c]{@{}c@{}}Motion\\ Seg.\end{tabular}}} \\
\midrule
\ch & \xm & \xm & 90.82 & 87.78 & 85.25 & - & - & - & - & - \\
\xm & \ch & \xm & - & - & - & 94.01 & \textbf{96.76} & 80.04 & 81.00 & - \\
\xm & \xm & \ch & - & - & - & - & - & - & - & 52.60 \\
\ch & \ch & \xm & 95.40 & 93.00 & 87.13 & 95.10 & 95.78 & 82.10 & 83.33 & - \\
\xm & \ch & \ch & - & - & - & 95.10 & 96.33 & 81.11 & 82.50 & 68.12 \\
\rowcolor{gray9} \ch & \ch & \ch & \textbf{96.34} & \textbf{94.00} & \textbf{90.01} & \textbf{96.18} & 96.12 & \textbf{84.50} & \textbf{81.09} & \textbf{73.14} \\ 
\bottomrule
\end{tabular}
\end{adjustbox}
\end{table}

The MTL model shows significant gains over STL across all tasks. For object detection, MTL improves easy, moderate, and hard AP by $\mathbf{5.52\%}$, $\mathbf{6.22\%}$, and $\mathbf{4.76\%}$, respectively. For semantic segmentation, improvements of 1-3\% IoU are obtained across all classes. Motion segmentation shows the most dramatic gains from MTL, with IoU increased by $\mathbf{20.5\%}$ absolute from $\mathbf{52.6}$ to $\mathbf{73.1}$. This highlights the synergies obtained when learning the tasks together. Since most moving objects are cars, the additional features learned by the other two tasks contribute to motion segmentation. Also, using an additional past frame contributes directly to better performance.

While there are other multi-task learning approaches for autonomous driving perception, we could not find prior work tackling these three tasks on KITTI and SemanticKITTI in a joint framework. The improvements are attributable to shared representations in the encoder, improved regularization, and the inductive bias from training signals for related tasks. As shown in Table~\ref{tab:table1-mtl-stl}, the multi-task approach outperforms single-task learning across object detection, semantic segmentation, and motion segmentation metrics. The only exception is road segmentation IoU, where STL has a slight $\mathbf{0.6\%}$ edge. A reasonable explanation for the consistent improvements from MTL is that the additional guidance from related semantic and motion features helps regularize the model and improve generalization. Our results validate multi-task learning as an effective paradigm for combined perception in autonomous driving.\par

\begin{figure*}[!t]
  \centering
\includegraphics[width=\textwidth,height=2.0\textwidth,keepaspectratio]{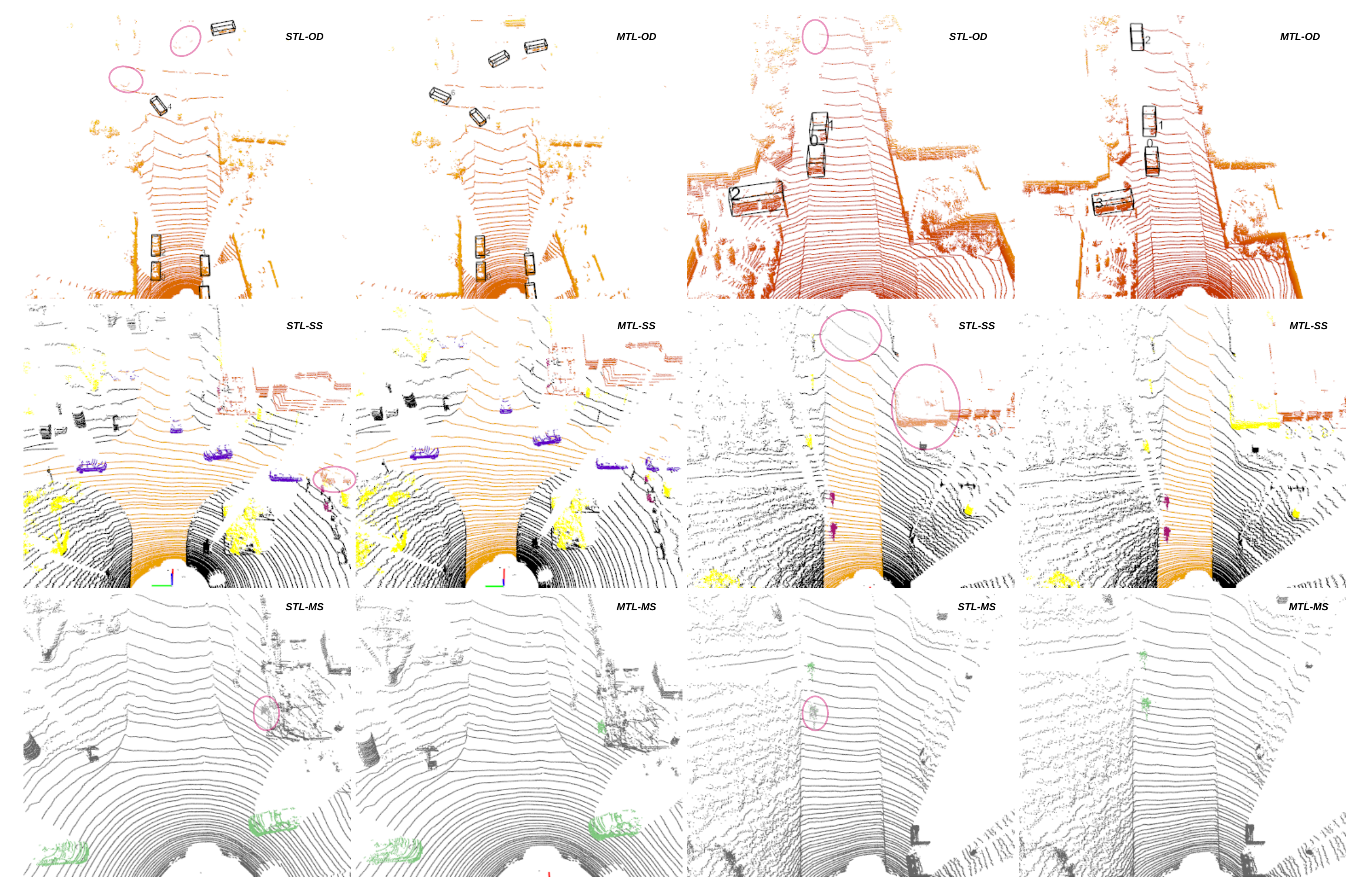}\hfill
\caption{\textbf{Qualitative results for performance comparison} - MTL and STL approaches for object detection, semantic segmentation, and motion segmentation. The MTL approach improves all tasks for the shown failure cases of STL. The results can be further viewed in high-quality in the video \url{https://youtu.be/H-hWRzv2lIY}.}
  \label{fig:qualitative_stl_mtl_comp}
\end{figure*}

We also compare our proposed LiDAR-BEVMTN against other LiDAR-only object detection methods on the KITTI BEV detection benchmark. Table~\ref{tab:table2-od} shows results using average precision (AP) at an IoU threshold of $\mathbf{0.5}$. 
Figure \ref{fig:qualitative_stl_mtl_comp} shows some of the failure cases of single-task networks, which are improved upon by the proposed multi-task approach.


LiDAR-BEVMTN achieves $\mathbf{96.34 AP}$ on the easy split, on par with top-performing SE-SSD at $\mathbf{97.7 AP}$. Our approach obtains state-of-the-art performance for the moderate category at $\mathbf{94.0 AP}$, a $\mathbf{0.7}$ point improvement over the prior best TED-S. While TED-S remains top for the hard category, LiDAR-BEVMTN attains a competitive $\mathbf{90.01 AP}$, just $\mathbf{2.1}$ points behind.\par

The strong performance of LiDAR-BEVMTN can be attributed to its multi-task learning framework. By training object detection jointly with semantic and motion segmentation in a shared model, the representation learns to encode spatial details, semantics, and motion cues. This provides a robust feature space for accurate 3D localization and orientation estimation. The results validate the benefits of multi-task inductive transfer for object detection from point cloud data.\par


\begin{table}[t]
\captionsetup{singlelinecheck=false, font=small, skip=2pt, belowskip=-2pt}
\centering
\caption{\textbf{BEV Object detection comparison} - AP at IoU=0.5 and inference latency in ms/frame against other single task approaches on Nvidia Xavier AGX. Other models were run on Nvidia without any optimizations.} 

\label{tab:table2-od}
\begin{adjustbox}{width=0.8\columnwidth}
\begin{tabular}{@{}lcccc@{}}
\toprule
\multicolumn{1}{c|}{} & \multicolumn{3}{c|}{\begin{tabular}[c]{@{}c@{}}Object Detection\\ IoU = 0.5\end{tabular}} &  \\ \cmidrule(lr){2-4}
\multicolumn{1}{c|}{\multirow{-2}{*}{\textit{Models}}} & \cellcolor[HTML]{9AFF99}Easy & \cellcolor[HTML]{FFFC9E}Moderate & \multicolumn{1}{c|}{\cellcolor[HTML]{FD6864}Hard} & \multirow{-2}{*}{\begin{tabular}[c]{@{}c@{}}Latency \\ (ms/frame)\end{tabular}} \\ \midrule
TED-S\cite{wu2023transformation} & 96.60 & 93.30 & \textbf{92.15} & 700 \\
SE-SSD\cite{zheng2021se} & \textbf{97.70} & 93.40 & 90.85 & 410 \\
Part-A2\cite{shi2019part} & 92.24 & 90.70 & 89.18 & 540 \\
PointRCNN\cite{shi2019pointrcnn} & 90.20 & 89.63 & 89.39 & 747 \\
\rowcolor {gray9} 
\textbf{LiDAR-BEVMTN(ours)} & 96.34 & \textbf{94.00} & 90.01 & \textbf{3.01} \\ \bottomrule
\end{tabular}
\end{adjustbox}
\end{table}
\begin{table}[t]
\captionsetup{singlelinecheck=false, font=small, skip=2pt, belowskip=-2pt}
\centering
\caption{\textbf{Intersection over Union (IoU) comparison for semantic segmentation on SemanticKITTI validation set}}
\label{tab:table-sem}
\begin{adjustbox}{width=0.8\columnwidth}
\small
\centering
\begin{tabular}{@{}l|cccc@{}}
\toprule
\multicolumn{1}{c|}{\multirow{2}{*}{\textit{Models}}} & 
\multicolumn{4}{c}{Semantic Segmentation - IoU} \\ \cmidrule(l){2-5} 
\multicolumn{1}{c|}{} &  
\cellcolor[HTML]{00b0f0}\textit{Car} &  
\cellcolor[HTML]{e5b9b5} \textit{Road} & 
\cellcolor[HTML]{a5a5a5} \textit{Building} &  
\cellcolor[HTML]{00b050} \textit{Veg.} \\
\midrule
Cylinder3D\cite{zhou2020cylinder3d}     & 95.35 & 93.36 & 90.33 & \textbf{88.37} \\
Point-Voxel-KD\cite{hou2022point} & {96.10} & 91.40 & \textbf{91.00} & 85.60 \\
SalsaNext\cite{cortinhal2020salsanext}      & 91.80 & 91.50 & 90.20 & 81.80 \\
RangeNet++\cite{milioto2019rangenet++}     & 91.40 & 91.80 & 87.40 & 80.50 \\
\rowcolor{gray9} \textbf{LiDAR-BEVMTN} & \textbf{96.18} & \textbf{96.12} & 84.50 & 81.09 \\ 
\bottomrule
\end{tabular}
\end{adjustbox}
\end{table}
Table~\ref{tab:table-sem} compares our LiDAR-BEVMTN to other LiDAR-based approaches on the SemanticKITTI validation set using BEV 2D intersection-over-union (IoU). LiDAR-BEVMTN achieves state-of-the-art performance for cars with $\mathbf{96.18\%}$ IoU, outperforming the next best Cylinder3D by $\mathbf{0.83\%}$. Our model also sets a new state-of-the-art for road segmentation at $\mathbf{96.12\%}$ IoU. LiDAR-BEVMTN obtains competitive performance for vegetation but lags behind Cylinder3D, by $\mathbf{7.28\%}$. On buildings, our approach attains $\mathbf{84.5\%}$ IoU, lower than Point-Voxel-KD and Cylinder3D, which incorporate structural priors. The lower performance on these classes can be attributed to the following factors - Both datasets present a significant bias towards cars and road as the most frequently occurring categories. Furthermore, our sampling data augmentation further increases the number of cars. Also, the SWAG module works in conjunction with the object detection head which is primarily detecting cars. Therefore, it enhances features for cars and roads (by spatial affinity), while not doing the same for buildings and vegetation. Another key factor is that the BEV representation presents advantage in regular grid based 2D representation of 3D structures. Classes like vegetation produce highly irregular and sparse point clouds, thereby not benefiting from this and making segmentation even harder.

The strong semantic segmentation capabilities of LiDAR-BEVMTN can be attributed to its multi-task learning framework. The model learns a spatially and contextually enriched representation by training jointly with related tasks like object detection. This provides a robust encoder space for accurate per-point semantic prediction. Our experiments validate multi-task learning as a practical paradigm for semantic segmentation from point cloud data.\par
\begin{table}[t]
\captionsetup{singlelinecheck=false, font=small, skip=2pt, belowskip=-2pt}
\centering
\caption{\textbf{Intersection over Union (IoU) comparison for moving object segmentation on the SemanticKITTI validation set (seq 08).} The number of past frames used by each method is shown under history.}
\label{tab:table4-mot}
\scalebox{0.6}{
\begin{adjustbox}{width=\columnwidth}
\small
\begin{tabular}{@{}l|cc@{}}
\toprule
\multicolumn{1}{c|}{\multirow{2}{*}{\textit{Models}}} 
& \multicolumn{2}{l}{\textit{Motion Segmentation}} \\ 
\cmidrule(l){2-3} 
\multicolumn{1}{c|}{}                                 
& \multicolumn{1}{c|}{History}        & IoU        \\ 
\midrule
RVMOS\cite{kim2022rvmos}          & \multicolumn{1}{c|} 5   & 71.20 \\
MotionSeg3D\cite{sun2022efficient}    & \multicolumn{1}{c|} 8   & 65.20 \\
4DMOS\cite{mersch2022receding}          & \multicolumn{1}{c|}{10} & 65.20 \\
LMNet\cite{chen2021moving}          & \multicolumn{1}{c|} 8   & 62.50 \\
LiMoSeg\cite{mohapatra2021limoseg}        & \multicolumn{1}{c|} 2   & 52.60 \\
\rowcolor{gray9}  \textbf{LiDAR-BEVMTN} & \multicolumn{1}{c|} 3  & \textbf{73.14} \\ 
\bottomrule
\end{tabular}
\end{adjustbox}
}
\end{table}
Table~\ref{tab:table4-mot} compares our LiDAR-BEVMTN to other LiDAR-based moving object segmentation methods on the SemanticKITTI validation set. Using just $\mathbf{3}$ previous frames, LiDAR-BEVMTN achieves state-of-the-art performance with $\mathbf{73.14\%}$ IoU, outperforming the next best RVMOS ($\mathbf{71.2\%}$ with $\mathbf{5}$ frames) that utilizes more history. Compared to other approaches using $\mathbf{8-10}$ past frames, our model with $\mathbf{3}$ frames still shows substantial gains of $\mathbf{7-10\%}$ IoU.\par
The model learns spatial and temporal features useful for tracking motion by training jointly with related perception tasks. This provides a robust encoder space for accurate per-point moving object prediction with minimal history. Our experiments validate the benefits of multi-task representation learning for motion segmentation from point cloud sequences.\par
\begin{table}[t]
\centering
\captionsetup{singlelinecheck=false, font=small, skip=2pt, belowskip=-4pt}
\caption{\textbf{Ablation: Effect of semantic guidance and range-based point cloud densification on different tasks.} Point cloud augmentation was not applied to segmentation tasks.}
\label{tab:my-table-abl-composite}
\begin{adjustbox}{width=\columnwidth}
\setlength{\tabcolsep}{0.2em}
\begin{tabular}{@{}cccccccccc@{}}
\toprule
\multicolumn{2}{c|}{Param.} & 
\multicolumn{3}{c|}{Object Detection} & 
\multicolumn{4}{c|}{Semantic Segmentation} &  \\ 
\cmidrule(lr){3-9}
SWAG & \multicolumn{1}{c|}{Densif.} & 
\textit{\cellcolor[HTML]{a5eb8d}Easy} & 
\textit{\cellcolor[HTML]{96bbce}Moderate} & 
\multicolumn{1}{c|}{\textit{\cellcolor[HTML]{fc4538}Hard}} & 
\cellcolor[HTML]{00b0f0}\textit{Car} & 
\cellcolor[HTML]{e5b9b5} \textit{Road} & 
\cellcolor[HTML]{a5a5a5} \textit{Building} & 
\multicolumn{1}{c|}{\cellcolor[HTML]{00b050} \textit{Veg.}} & 
\multirow{-2}{*}{\begin{tabular}[c]{@{}c@{}}Motion\\ Seg.\end{tabular}} \\ 
\midrule
\xm & \xm & 93.01 & 90.49 & 87.04 & 94.14 & 94.26 & 83.22 & \textbf{81.49} & 71.12 \\
\xm & \ch & 94.81 & 92.19 & 88.10 & NA & NA & NA & NA & NA \\
\ch & \xm & 93.50 & 91.34 & 88.12 & 96.15 & 96.12 & 84.50 & 81.10 & 73.01 \\
\rowcolor{gray9}\ch & \ch & \textbf{96.34} & \textbf{94.00} & \textbf{90.01} & \textbf{96.18} & \textbf{96.12} & \textbf{84.50} & 81.09 & \textbf{73.14} \\
\bottomrule
\end{tabular}
\end{adjustbox}
\end{table}
Our ablation studies in Table~\ref{tab:my-table-abl-composite} analyze the impact of the proposed SWAG module and point cloud densification on object detection, semantic segmentation, and motion segmentation. Using SWAG improves object detection AP in easy, moderate, and hard categories by $\mathbf{0.5-1.4\%}$ absolute. This supports our hypothesis that incorporating weighted semantic features benefits object detection. SWAG also provides small IoU gains of $\mathbf{0.5-1.3\%}$ for semantic segmentation across cars, roads, and buildings. SWAG consistently improves the utilization of semantic features, benefiting object detection and semantic segmentation. Point cloud densification shows marginal AP improvements of $\mathbf{0.8-1.9\%}$ in moderate and hard object detection categories, at the cost of a $\mathbf{1.2\%}$ drop in easy. This mixed impact is likely because densification was only applied to object detection inputs during multi-task training. The variation in BEV input causes negative effects on joint learning. Point cloud densification shows potential but needs a unified augmentation strategy across tasks.\par

We also train and evaluate our method on the Waymo dataset \cite{sun2020scalability} and compare the results with another multi-task LiDAR network\cite{ye2022lidarmultinet}. There is limited work in LiDAR multi-task models which makes further comparision difficult. Table \ref{tab:waymo-val-od-semseg} shows the quantitative results for object detection and semantic segmentation.
\\
For object detection, we follow \cite{ye2022lidarmultinet} and use 3 past frames. For vehicle class we achieve an accuracy very close to \cite{ye2022lidarmultinet}. For pedestrian and cyclist classes, there is a significant difference in accuracy. It is likely because BEV is not a suitable representation for objects with a small cross section in x-y plane which occupy very few pixels (\(<\) 5) in BEV space. This makes detecting pedestrians, localizing the center and learning orientation very difficult, leading to lower IoU. Furthermore, since BEV collapses z-dimension, points along the height of an object at the same x-y location do not add more information. However, our BEV based model has siginificantly better computational efficiency compared to \cite{ye2022lidarmultinet}.

We have introduced a new data pre-processing technique to overcome this to some extent. We divide the expected z-axis range into 21 bins of equal size and create 21 channels in the BEV accordingly. This allows us to represent upto 21 3D points per pixel in BEV where the points map to the same pixel coordinates in x-y plane. This was found to improve detection performance in our experiments.
Also, \cite{ye2022lidarmultinet} uses 3D convolutions in it's 3D encoder-decoder architecture whereas we use purely 2D convolutions for runtime optimization.
\\
For semantic segmentation, we achieve the best IoU for car and road classes since they constitute the majority of the semantic labels. Furthermore, their larger cross section allows better classification. We could not find class wise segmentation scores in \cite{ye2022lidarmultinet} for Waymo validation set. Conversion from BEV space to 3D space for benchmarking has some loss in score.

\begin{table}[t]
\centering
\captionsetup{singlelinecheck=false, font=small, skip=2pt, belowskip=-2pt}
\caption{\textbf{Inference latency comparison}- MTL and STL networks for object detection, semantic segmentation, and motion segmentation. MTL - 3 tasks achieves 3x speed up over three single task networks.}
\label{tab:table-latency-comparison}
\begin{adjustbox}{width=\columnwidth}
\setlength{\tabcolsep}{0.2em}
\begin{tabular}{@{}l|ccc|c@{}}
\toprule
\multicolumn{1}{c|}{} & 
\multicolumn{3}{c|}{FP16 inference latency (ms)} & \\ 
\cmidrule(lr){2-4}
\multicolumn{1}{c|}{\multirow{-2}{*}{Task combination}} &
  \cellcolor[HTML]{FFCCC9}\textit{Encoder} &
  \cellcolor[HTML]{FFFFC7}\textit{\begin{tabular}[c]{@{}c@{}}Decoder+\\ Heads\end{tabular}} &
  \cellcolor[HTML]{ECF4FF}\textit{Total} &
  \multirow{-2}{*}{\begin{tabular}[c]{@{}c@{}}Speedup\\ factor\end{tabular}} \\ \midrule
STL(OD) + STL(SS) + STL(MS) & \multicolumn{1}{c|}{6.80}          & \multicolumn{1}{c|}{2.00}          & 8.80          & baseline      \\
STL (OD) + MTL(SS + MS)     & \multicolumn{1}{c|}{4.00}          & \multicolumn{1}{c|}{1.65}          & 5.65          & 1.55          \\
STL(MS) + MTL(OD + SS)      & \multicolumn{1}{c|}{4.00}          & \multicolumn{1}{c|}{1.83}          & 5.83          & 1.50          \\
\textbf{MTL (OD + SS + MS)} & \multicolumn{1}{c|}{\textbf{2.00}} & \multicolumn{1}{c|}{\textbf{1.01}} & \textbf{3.01} & \textbf{2.92} \\ \bottomrule
\end{tabular}
\end{adjustbox}
\end{table}

\begin{table}[t]
\centering
\captionsetup{singlelinecheck=false, font=small, skip=2pt, belowskip=-2pt}
\caption{\textbf{Comparison between static data augmentations vs progressive data augmentation.}}
\label{tab:data-aug-static-progressive}
\begin{adjustbox}{width=\columnwidth}
\setlength{\tabcolsep}{0.2em}
\begin{tabular}{@{}l|lllllll|l@{}}
\toprule
 & \multicolumn{3}{c|}{Object Detection} & \multicolumn{4}{c|}{Semantic Segmentation} & \multicolumn{1}{c}{} \\ \cmidrule(lr){2-8}
\multirow{-2}{*}{Augmentation} & \multicolumn{1}{c}{\cellcolor[HTML]{9AFF99}Easy} & \multicolumn{1}{c}{\cellcolor[HTML]{FFFFC7}Mod.} & \multicolumn{1}{c|}{\cellcolor[HTML]{FFCCC9}Hard} & \multicolumn{1}{c}{\cellcolor[HTML]{96FFFB}Car} & \multicolumn{1}{c}{\cellcolor[HTML]{FFCCC9}Road} & \multicolumn{1}{c}{\cellcolor[HTML]{EFEFEF}Build.} & \multicolumn{1}{c|}{\cellcolor[HTML]{32CB00}Veg.} & \multicolumn{1}{c}{\multirow{-2}{*}{\begin{tabular}[c]{@{}c@{}}Motion\\ Seg.\end{tabular}}} \\ \midrule
Static aug. & 96.27 & 94.20 & 89.10 & 93.00 & 95.32 & 84.35 & 81.10 & 68.16 \\
\textbf{Progressive aug.} & \textbf{96.34} & 94.00 & \textbf{90.01} & \textbf{96.18} & \textbf{96.12} & \textbf{84.50} & 81.09 & \textbf{73.14} \\ \bottomrule
\end{tabular}
\end{adjustbox}
\end{table}


\begin{table}[t!]
\centering
\captionsetup{singlelinecheck=false, font=small, skip=2pt, belowskip=-2pt}
\caption{\textbf{Runtime analysis of encoder and decoder building blocks
under different optimization conditions in a single task context:} vanilla pytorch (pth), pruning with maximum 50\% sparsity (prune), conversion to tensorrt FP16 (trt16), pruning followed by conversion to tensorrt FP16 (prune + trt16)}
\label{tab:timing-details}
\scalebox{0.99}{
\begin{adjustbox}{width=\columnwidth}
\begin{tabular}{@{}c|cccccccc@{}}
\toprule
 & \multicolumn{4}{c|}{\begin{tabular}[c]{@{}c@{}}Downsampling Blocks\\ (Encoder)\end{tabular}} & \multicolumn{4}{c}{\begin{tabular}[c]{@{}c@{}}Upsampling Blocks\\ (Decoder)\end{tabular}} \\ \cmidrule(l){2-9} 
\multirow{-2}{*}{\begin{tabular}[c]{@{}c@{}}Block\\ index\end{tabular}} & \cellcolor[HTML]{FFCCC9}\textit{\textbf{pth}} & \cellcolor[HTML]{FFCE93}\textit{\textbf{prune}} & \cellcolor[HTML]{FFFC9E}\textit{\textbf{trt16}} & \multicolumn{1}{c|}{\cellcolor[HTML]{9AFF99}\textit{\textbf{\begin{tabular}[c]{@{}c@{}}prune +\\ trt16\end{tabular}}}} & \cellcolor[HTML]{FFCCC9}\textit{\textbf{pth}} & \cellcolor[HTML]{FFCE93}\textit{\textbf{prune}} & \cellcolor[HTML]{FFFC9E}\textit{\textbf{trt16}} & \cellcolor[HTML]{9AFF99}\textit{\textbf{\begin{tabular}[c]{@{}c@{}}prune +\\ trt16\end{tabular}}} \\ \midrule
1 & 20.99 & 12.10 & 4.11 & 2.12 & 11.20 & 8.10 & 2.16 & 1.02 \\
2 & 11.81 & 8.03 & 2.12 & 1.48 & 7.69 & 5.11 & 1.00 & 0.40 \\
3 & 7.98 & 4.45 & 1.20 & 1.30 & 6.11 & 4.71 & 0.56 & 0.30 \\
4 & 6.98 & 3.01 & 1.10 & 1.05 & 5.57 & 3.96 & 0.61 & 0.23 \\
5 & 6.19 & 3.01 & 1.10 & 1.01 & 5.07 & 2.96 & 0.50 & 0.12 \\
\rowcolor[HTML]{EFEFEF} 
\textit{Total} & 53.95 & 30.60 & 9.63 & \textbf{6.96} & 35.64 & 24.84 & 4.83 & \textbf{2.07} \\ \bottomrule
\end{tabular}
\end{adjustbox}
}
\end{table}

\begin{table*}[htb!]
\caption{\textbf{Evaluation on Waymo validation set:} Comparision of 3D object detection and semantic segmentation on the Waymo validation set against another multi-task lidar only method.
Class-wise IoU was not available for LidarMultiNet for semantic segmentation on the validation set.}
\label{tab:waymo-val-od-semseg}
\begin{tabular}{@{}c|ccc|cccccccccc@{}}
\toprule
 & \multicolumn{3}{c|}{\begin{tabular}[c]{@{}c@{}}Object detection\\ mAPH - L2\end{tabular}} & \multicolumn{10}{c}{\begin{tabular}[c]{@{}c@{}}Semantic segmentation \\ class wise IoU and mIoU\end{tabular}} \\ \cmidrule(l){2-14} 
\multirow{-2}{*}{Method} & \cellcolor[HTML]{FFCCC9}\textit{Veh.} & \cellcolor[HTML]{F1B3F7}\textit{Ped.} & \cellcolor[HTML]{ECF4FF}\textit{Cyc.} & \cellcolor[HTML]{FFCCC9}\textit{Car} & \cellcolor[HTML]{FFCE93}\textit{Truck} & \cellcolor[HTML]{FFFC9E}\textit{Bus} & \cellcolor[HTML]{F1B3F7}\textit{Ped.} & \cellcolor[HTML]{ECF4FF}\textit{Bicyclist} & \cellcolor[HTML]{CBCEFB}\textit{Motorcyclist} & \cellcolor[HTML]{9AFF99}\textit{Veg.} & \cellcolor[HTML]{96FFFB}\textit{Building} & \cellcolor[HTML]{C0C0C0}\textit{Road} & \textbf{mIoU} \\ \midrule
LidarMultiNet\cite{ye2022lidarmultinet} & \textbf{73.38} & \textbf{74.10} & \textbf{77.96} & - & - & - & - & - & - & - & - & - & 72.40 \\
\cellcolor[HTML]{EFEFEF}\textbf{LiDAR-BEVMTN} & 72.70 & 69.42 & 71.10 & \cellcolor[HTML]{EFEFEF}89.77 & \cellcolor[HTML]{EFEFEF}60.92 & \cellcolor[HTML]{EFEFEF}65.12 & \cellcolor[HTML]{EFEFEF}60.64 & \cellcolor[HTML]{EFEFEF}61.32 & \cellcolor[HTML]{EFEFEF}61.01 & \cellcolor[HTML]{EFEFEF}88.63 & \cellcolor[HTML]{EFEFEF}80.13 & \cellcolor[HTML]{EFEFEF}93.10 & \cellcolor[HTML]{EFEFEF}\textbf{73.40} \\ \bottomrule
\end{tabular}
\end{table*}
\par

Table~\ref{tab:table-latency-comparison} compares the inference latency of our multi-task LiDAR-BEVMTN against single-task LiDAR networks for object detection, semantic segmentation, and motion segmentation. Using three separate single-task models for the three tasks takes $\mathbf{8.80ms}$ per frame. Combining two tasks under multi-task learning reduces latency to $\mathbf{5.65-5.83ms}$, a $\mathbf{1.5-1.55x}$ speedup. Finally, with our full multi-task model concurrently performing object detection, semantic segmentation, and motion segmentation, inference time is reduced to just $\mathbf{3.01ms}$ per frame - a $\mathbf{2.92x}$ speedup over separate single-task networks. \par 
The substantial latency improvements of LiDAR-BEVMTN can be attributed to computation sharing in the unified encoder-decoder network. Features are extracted just once from the LiDAR point cloud and reused across all heads. This avoids expensive redundant feature extraction for each task. Our experiments validate multi-task learning as an effective approach to improve computational efficiency and enable real-time performance for combined 3D perception from point cloud data.

\textbf{Timing analysis and optimization:}
Table \ref{tab:timing-details} presents a microscopic analysis of inference latency of the building blocks of the encoder and decoder in a single task context. The block indices refer to the 5-stage encoder and decoder hierarchically as shown in Figure \ref{fig:Figure_overall_nw_bd}. While the decoder is implemented as a monolithic architecture, we time each block separately along the longest chain for this detailed analysis. The encoder and decoder and significantly different in architecture. The encoder is a sequential architecture on a block level, i.e., each DB stage waits on results from it's higher stage to start executing. Spatial resolution of input reduces by half at each DB while channels increase as 32-64-128-256-512 respectively. Reduction is spatial dimension translates directly to reduction in runtime. This is because all operations along channels for a convolution can be parallelized. This is evident from the pytorch (pth) runtimes of DB blocks. We then apply weight pruning\cite{nni2021} with a maximum sparsity of 50\% as an optimization technique. The early stages of DB see the most sparse input since the input BEV is sparse and produce sparse features. This results in maximum pruning in the early DB stages. As seen in the table, deeper stages of DB produce progressively denser feature maps, thereby allowing lesser pruning. This results in lesser gains in runtime reduction as depth of DB increases. Finally we compile the model into a tensorrt engine at FP16 precision to achieve the lowest inference latency. 
\\
The decoder is parallel on multiple levels. 
Along each column in the decoder, all UBs can execute parallelly since all necessary inputs for each block would be already available at the end of execution of the previous column.
For example - the first UB needs f1 and f2 inputs (from encoder) to start executing. The arrow from the top of UB designates a bypass of it's input to output. 
Internally each UB is much thinner than a DB and therefore UBs execute significantly faster than their corresponding DBs as seen in the table. Furthermore, reduction in spatial dimensions translates to faster runtime as is the case with UBs. Since different UBs in a column have different execution time, UBs in successive columns can start executing as soon as their inputs are ready from the previous column of UBs. Another feature of the decoder is that the number of UBs competing for resources reduces linearly along rows and columns. This leads to significant reduction in runtime of the inner layers of the decoder and allows successive blocks in the outer layer to run progressively faster. \\
We arrive at the following general optimization principles from our analysis:
\begin{itemize}
    \item Input size impacts latency directly - smaller size BEV leads to faster inference speeds and lesser sparsity
    \item Inference latency increases with increase in depth of the network
    \item Convolutions are more efficient along the channel dimension - increasing channels while reducing feature map sizes leads to faster inference times.
    \item Restricting maximum size along all dimensions to 1024 or 512 is advantageous for parallelization with CUDA (underlying framework on Nvidia GPUs)
    \item Convolution layers with multiple of 32 kernels are advantageous for maximum number of concurrent warp execution on Nvidia GPUs
    \item Pruning at a higher maximum sparsity allows better optimization - maximum sparsity only imposes an upper bound, individual layers may have different amount of sparsity
    \item Compiling with Tensorrt at FP16 quantization provides the most significant boost in inference speed.
    \item Network architecture with maximum number of independently executable modules achieves better inference speed following the principles of latency-hiding and parallel execution.
\end{itemize}
Finally, we also observe that whole network optimization yields slightly faster inference speed that block level optimization and sum.



\section{Conclusion}

In this work, we have presented LiDAR-BEVMTN, a multi-task neural network for joint object detection, semantic segmentation, and motion segmentation from LiDAR point clouds. Our proposed model achieves state-of-the-art results across all three tasks on the challenging KITTI and SemanticKITTI datasets. To the best of our knowledge, this is the first work to tackle these specific tasks together using multi-task deep learning on KITTI and SemanticKITTI. Our results and analyses demonstrate the advantages of multi-task learning for combined 3D perception. Despite multiple tasks using consecutive LiDAR scans, we achieved a low latency of 3 ms on an embedded automotive platform. Future work includes extending this approach to other perception tasks like odometry, ego-motion estimation, and velocity prediction for autonomous driving. We hope this work provides useful insights and motivates continued research into multi-task deep learning for LiDAR-based scene understanding.
\bibliographystyle{IEEEtran}
\bibliography{bib/references}

\begin{thebibliography}{10}
\providecommand{\url}[1]{#1}
\csname url@samestyle\endcsname
\providecommand{\newblock}{\relax}
\providecommand{\bibinfo}[2]{#2}
\providecommand{\BIBentrySTDinterwordspacing}{\spaceskip=0pt\relax}
\providecommand{\BIBentryALTinterwordstretchfactor}{4}
\providecommand{\BIBentryALTinterwordspacing}{\spaceskip=\fontdimen2\font plus
\BIBentryALTinterwordstretchfactor\fontdimen3\font minus \fontdimen4\font\relax}
\providecommand{\BIBforeignlanguage}[2]{{%
\expandafter\ifx\csname l@#1\endcsname\relax
\typeout{** WARNING: IEEEtran.bst: No hyphenation pattern has been}%
\typeout{** loaded for the language `#1'. Using the pattern for}%
\typeout{** the default language instead.}%
\else
\language=\csname l@#1\endcsname
\fi
#2}}
\providecommand{\BIBdecl}{\relax}
\BIBdecl

\bibitem{joseph2021autonomous}
L.~Joseph and A.~K. Mondal, \emph{{Autonomous driving and Advanced Driver-Assistance Systems (ADAS): applications, development, legal issues, and testing}}.\hskip 1em plus 0.5em minus 0.4em\relax CRC Press, 2021.

\bibitem{kumar2018near}
V.~R. Kumar, S.~Milz, C.~Witt, M.~Simon \emph{et~al.}, ``{Near-field depth estimation using monocular fisheye camera: A semi-supervised learning approach using sparse LiDAR data},'' in \emph{CVPR Workshop}, vol.~7, 2018, p.~2.

\bibitem{kumar2021svdistnet}
V.~R. Kumar, M.~Klingner, S.~Yogamani, M.~Bach \emph{et~al.}, ``{SVDistNet: Self-supervised near-field distance estimation on surround view fisheye cameras},'' \emph{IEEE Transactions on Intelligent Transportation Systems}, vol.~23, no.~8, pp. 10\,252--10\,261, 2021.

\bibitem{uricar2019challenges}
M.~Uric{\'a}r, D.~Hurych, P.~Krizek, and S.~Yogamani, ``{Challenges in designing datasets and validation for autonomous driving},'' \emph{arXiv preprint arXiv:1901.09270}, 2019.

\bibitem{uricar2019desoiling}
M.~Uric{\'a}r, J.~Ulicny, G.~Sistu, H.~Rashed \emph{et~al.}, ``{Desoiling dataset: Restoring soiled areas on automotive fisheye cameras},'' in \emph{Proceedings of the IEEE/CVF International Conference on Computer Vision Workshops}, 2019.

\bibitem{dhananjaya2021weather}
M.~M. Dhananjaya, V.~R. Kumar, and S.~Yogamani, ``{Weather and light level classification for autonomous driving: Dataset, baseline and active learning},'' in \emph{IEEE International Intelligent Transportation Systems Conference (ITSC)}.\hskip 1em plus 0.5em minus 0.4em\relax IEEE, 2021, pp. 2816--2821.

\bibitem{geiger2013vision}
A.~Geiger, P.~Lenz, C.~Stiller, and R.~Urtasun, ``{Vision meets robotics: The kitti dataset},'' \emph{The International Journal of Robotics Research}, vol.~32, no.~11, pp. 1231--1237, 2013.

\bibitem{behley2019semantickitti}
J.~Behley, M.~Garbade, A.~Milioto, J.~Quenzel \emph{et~al.}, ``{Semantickitti: A dataset for semantic scene understanding of lidar sequences},'' in \emph{Proceedings of the IEEE/CVF International Conference on Computer Vision}, 2019, pp. 9297--9307.

\bibitem{sun2020scalability}
P.~Sun, H.~Kretzschmar, X.~Dotiwalla, A.~Chouard \emph{et~al.}, ``{Scalability in perception for autonomous driving: Waymo open dataset},'' in \emph{Proceedings of the IEEE/CVF conference on computer vision and pattern recognition}, 2020, pp. 2446--2454.

\bibitem{ye2022lidarmultinet}
D.~Ye, Z.~Zhou, W.~Chen, Y.~Xie \emph{et~al.}, ``{Lidarmultinet: Towards a unified multi-task network for lidar perception},'' \emph{arXiv preprint arXiv:2209.09385}, 2022.

\bibitem{zhou2018voxelnet}
Y.~Zhou and O.~Tuzel, ``{Voxelnet: End-to-end learning for point cloud based 3d object detection},'' in \emph{Proceedings of the IEEE conference on computer vision and pattern recognition}, 2018, pp. 4490--4499.

\bibitem{maturana2015voxnet}
D.~Maturana and S.~Scherer, ``{Voxnet: A 3d convolutional neural network for real-time object recognition},'' in \emph{IEEE/RSJ international conference on intelligent robots and systems (IROS)}.\hskip 1em plus 0.5em minus 0.4em\relax IEEE, 2015, pp. 922--928.

\bibitem{yan2018second}
Y.~Yan, Y.~Mao, and B.~Li, ``{Second: Sparsely embedded convolutional detection},'' \emph{Sensors}, vol.~18, no.~10, p. 3337, 2018.

\bibitem{shi2019pointrcnn}
S.~Shi, X.~Wang, and H.~Li, ``{Pointrcnn: 3d object proposal generation and detection from point cloud},'' in \emph{Proceedings of the IEEE/CVF conference on computer vision and pattern recognition}, 2019, pp. 770--779.

\bibitem{qi2017pointnet++}
C.~R. Qi, L.~Yi, H.~Su, and L.~J. Guibas, ``{Pointnet++: Deep hierarchical feature learning on point sets in a metric space},'' \emph{Advances in neural information processing systems}, vol.~30, 2017.

\bibitem{shi2020pv}
S.~Shi, C.~Guo, L.~Jiang, Z.~Wang \emph{et~al.}, ``{Pv-rcnn: Point-voxel feature set abstraction for 3d object detection},'' in \emph{Proceedings of the IEEE/CVF Conference on Computer Vision and Pattern Recognition}, 2020, pp. 10\,529--10\,538.

\bibitem{wu2022sparse}
X.~Wu, L.~Peng, H.~Yang, L.~Xie \emph{et~al.}, ``{Sparse fuse dense: Towards high quality 3d detection with depth completion},'' in \emph{Proceedings of the IEEE/CVF Conference on Computer Vision and Pattern Recognition}, 2022, pp. 5418--5427.

\bibitem{borse2023x}
S.~Borse, M.~Klingner, V.~R. Kumar, H.~Cai \emph{et~al.}, ``{X-Align: Cross-Modal Cross-View Alignment for Bird's-Eye-View Segmentation},'' in \emph{Proceedings of the IEEE/CVF Winter Conference on Applications of Computer Vision}, 2023, pp. 3287--3297.

\bibitem{mohapatra2021bevdetnet}
S.~Mohapatra, S.~Yogamani, H.~Gotzig, S.~Milz \emph{et~al.}, ``{BEVDetNet: bird's eye view LiDAR point cloud based real-time 3D object detection for autonomous driving},'' in \emph{IEEE International Intelligent Transportation Systems Conference (ITSC)}.\hskip 1em plus 0.5em minus 0.4em\relax IEEE, 2021, pp. 2809--2815.

\bibitem{yang2018pixor}
B.~Yang, W.~Luo, and R.~Urtasun, ``{Pixor: Real-time 3d object detection from point clouds},'' in \emph{Proceedings of the IEEE conference on Computer Vision and Pattern Recognition}, 2018, pp. 7652--7660.

\bibitem{simony2018complex}
M.~Simony, S.~Milzy, K.~Amendey, and H.-M. Gross, ``{Complex-yolo: An euler-region-proposal for real-time 3d object detection on point clouds},'' in \emph{Proceedings of the European Conference on Computer Vision (ECCV) Workshops}, 2018.

\bibitem{lang2019pointpillars}
A.~H. Lang, S.~Vora, H.~Caesar, L.~Zhou \emph{et~al.}, ``{Pointpillars: Fast encoders for object detection from point clouds},'' in \emph{Proceedings of the IEEE/CVF conference on computer vision and pattern recognition}, 2019, pp. 12\,697--12\,705.

\bibitem{liu2016ssd}
W.~Liu, D.~Anguelov, D.~Erhan, C.~Szegedy \emph{et~al.}, ``{Ssd: Single shot multibox detector},'' in \emph{European conference on computer vision}.\hskip 1em plus 0.5em minus 0.4em\relax Springer, 2016, pp. 21--37.

\bibitem{zheng2021se}
W.~Zheng, W.~Tang, L.~Jiang, and C.-W. Fu, ``{SE-SSD: Self-ensembling single-stage object detector from point cloud},'' in \emph{Proceedings of the IEEE/CVF Conference on Computer Vision and Pattern Recognition}, 2021, pp. 14\,494--14\,503.

\bibitem{hoang2024tsstdet}
H.~A. Hoang, D.~C. Bui, and M.~Yoo, ``Tsstdet: Transformation-based 3-d object detection via a spatial shape transformer,'' \emph{IEEE Sensors Journal}, 2024.

\bibitem{dosovitskiy2020image}
A.~Dosovitskiy, L.~Beyer, A.~Kolesnikov, D.~Weissenborn, X.~Zhai, T.~Unterthiner, M.~Dehghani, M.~Minderer, G.~Heigold, S.~Gelly \emph{et~al.}, ``An image is worth 16x16 words: Transformers for image recognition at scale,'' \emph{arXiv preprint arXiv:2010.11929}, 2020.

\bibitem{wu2023virtual}
H.~Wu, C.~Wen, S.~Shi, X.~Li, and C.~Wang, ``Virtual sparse convolution for multimodal 3d object detection,'' in \emph{Proceedings of the IEEE/CVF Conference on Computer Vision and Pattern Recognition}, 2023, pp. 21\,653--21\,662.

\bibitem{li2023logonet}
X.~Li, T.~Ma, Y.~Hou, B.~Shi, Y.~Yang, Y.~Liu, X.~Wu, Q.~Chen, Y.~Li, Y.~Qiao \emph{et~al.}, ``Logonet: Towards accurate 3d object detection with local-to-global cross-modal fusion,'' in \emph{Proceedings of the IEEE/CVF Conference on Computer Vision and Pattern Recognition}, 2023, pp. 17\,524--17\,534.

\bibitem{dutta2022vit}
P.~Dutta, G.~Sistu, S.~Yogamani, E.~Galv{\'a}n \emph{et~al.}, ``{ViT-BEVSeg: A hierarchical transformer network for monocular birds-eye-view segmentation},'' in \emph{International Joint Conference on Neural Networks (IJCNN)}.\hskip 1em plus 0.5em minus 0.4em\relax IEEE, 2022, pp. 1--7.

\bibitem{rashed2019motion}
H.~Rashed, A.~El~Sallab, S.~Yogamani, and M.~ElHelw, ``{Motion and depth augmented semantic segmentation for autonomous navigation},'' in \emph{Proceedings of the IEEE/CVF Conference on Computer Vision and Pattern Recognition Workshops}, 2019.

\bibitem{wu2018squeezeseg}
B.~Wu, A.~Wan, X.~Yue, and K.~Keutzer, ``{Squeezeseg: Convolutional neural nets with recurrent crf for real-time road-object segmentation from 3d lidar point cloud},'' in \emph{IEEE International Conference on Robotics and Automation (ICRA)}.\hskip 1em plus 0.5em minus 0.4em\relax IEEE, 2018, pp. 1887--1893.

\bibitem{wu2019squeezesegv2}
B.~Wu, X.~Zhou, S.~Zhao, X.~Yue \emph{et~al.}, ``{Squeezesegv2: Improved model structure and unsupervised domain adaptation for road-object segmentation from a lidar point cloud},'' in \emph{International Conference on Robotics and Automation (ICRA)}.\hskip 1em plus 0.5em minus 0.4em\relax IEEE, 2019, pp. 4376--4382.

\bibitem{iandola2016squeezenet}
F.~N. Iandola, S.~Han, M.~W. Moskewicz, K.~Ashraf \emph{et~al.}, ``{SqueezeNet: AlexNet-level accuracy with 50x fewer parameters and< 0.5 MB model size},'' \emph{arXiv preprint arXiv:1602.07360}, 2016.

\bibitem{zheng2015conditional}
S.~Zheng, S.~Jayasumana, B.~Romera-Paredes, V.~Vineet \emph{et~al.}, ``{Conditional random fields as recurrent neural networks},'' in \emph{Proceedings of the IEEE international conference on computer vision}, 2015, pp. 1529--1537.

\bibitem{milioto2019rangenet++}
A.~Milioto, I.~Vizzo, J.~Behley, and C.~Stachniss, ``{Rangenet++: Fast and accurate lidar semantic segmentation},'' in \emph{IEEE/RSJ international conference on intelligent robots and systems (IROS)}.\hskip 1em plus 0.5em minus 0.4em\relax IEEE, 2019, pp. 4213--4220.

\bibitem{redmon2018yolov3}
J.~Redmon and A.~Farhadi, ``{Yolov3: An incremental improvement},'' \emph{arXiv preprint arXiv:1804.02767}, 2018.

\bibitem{hou2022point}
Y.~Hou, X.~Zhu, Y.~Ma, C.~C. Loy \emph{et~al.}, ``{Point-to-voxel knowledge distillation for lidar semantic segmentation},'' in \emph{Proceedings of the IEEE/CVF Conference on Computer Vision and Pattern Recognition}, 2022, pp. 8479--8488.

\bibitem{kong2023lasermix}
L.~Kong, J.~Ren, L.~Pan, and Z.~Liu, ``Lasermix for semi-supervised lidar semantic segmentation,'' in \emph{Proceedings of the IEEE/CVF Conference on Computer Vision and Pattern Recognition}, 2023, pp. 21\,705--21\,715.

\bibitem{zhang2023growsp}
Z.~Zhang, B.~Yang, B.~Wang, and B.~Li, ``Growsp: Unsupervised semantic segmentation of 3d point clouds,'' in \emph{Proceedings of the IEEE/CVF Conference on Computer Vision and Pattern Recognition}, 2023, pp. 17\,619--17\,629.

\bibitem{li2024tfnet}
R.~Li, S.~Li, X.~Chen, T.~Ma, J.~Gall, and J.~Liang, ``Tfnet: Exploiting temporal cues for fast and accurate lidar semantic segmentation,'' in \emph{Proceedings of the IEEE/CVF Conference on Computer Vision and Pattern Recognition}, 2024, pp. 4547--4556.

\bibitem{chen2021moving}
X.~Chen, S.~Li, B.~Mersch, L.~Wiesmann \emph{et~al.}, ``{Moving object segmentation in 3D LiDAR data: A learning-based approach exploiting sequential data},'' \emph{IEEE Robotics and Automation Letters}, vol.~6, no.~4, pp. 6529--6536, 2021.

\bibitem{mohapatra2021limoseg}
S.~Mohapatra, M.~Hodaei, S.~Yogamani, S.~Milz \emph{et~al.}, ``{LiMoSeg: Real-time Bird's Eye View based LiDAR Motion Segmentation},'' \emph{arXiv preprint arXiv:2111.04875}, 2021.

\bibitem{sun2022efficient}
J.~Sun, Y.~Dai, X.~Zhang, J.~Xu \emph{et~al.}, ``{Efficient Spatial-Temporal Information Fusion for LiDAR-Based 3D Moving Object Segmentation},'' in \emph{2022 IEEE/RSJ International Conference on Intelligent Robots and Systems (IROS)}.\hskip 1em plus 0.5em minus 0.4em\relax IEEE, 2022, pp. 11\,456--11\,463.

\bibitem{zeng2024mambamos}
K.~Zeng, H.~Shi, J.~Lin, S.~Li, J.~Cheng, K.~Wang, Z.~Li, and K.~Yang, ``Mambamos: Lidar-based 3d moving object segmentation with motion-aware state space model,'' in \emph{Proceedings of the 32nd ACM International Conference on Multimedia}, 2024, pp. 1505--1513.

\bibitem{leang2020dynamic}
I.~Leang, G.~Sistu, F.~B{\"u}rger, A.~Bursuc \emph{et~al.}, ``{Dynamic task weighting methods for multi-task networks in autonomous driving systems},'' in \emph{IEEE 23rd International Conference on Intelligent Transportation Systems (ITSC)}.\hskip 1em plus 0.5em minus 0.4em\relax IEEE, 2020, pp. 1--8.

\bibitem{klingner2022detecting}
M.~Klingner, V.~R. Kumar, S.~Yogamani, A.~B{\"a}r \emph{et~al.}, ``{Detecting adversarial perturbations in multi-task perception},'' in \emph{IEEE/RSJ International Conference on Intelligent Robots and Systems (IROS)}.\hskip 1em plus 0.5em minus 0.4em\relax IEEE, 2022, pp. 13\,050--13\,057.

\bibitem{liu2022bevfusion}
Z.~Liu, H.~Tang, A.~Amini, X.~Yang \emph{et~al.}, ``{BEVFusion: Multi-Task Multi-Sensor Fusion with Unified Bird's-Eye View Representation},'' \emph{arXiv preprint arXiv:2205.13542}, 2022.

\bibitem{duffhauss2020pillarflownet}
F.~Duffhauss and S.~A. Baur, ``{PillarFlowNet: A real-time deep multitask network for LiDAR-based 3D object detection and scene flow estimation},'' in \emph{IEEE/RSJ International Conference on Intelligent Robots and Systems (IROS)}.\hskip 1em plus 0.5em minus 0.4em\relax IEEE, 2020, pp. 10\,734--10\,741.

\bibitem{feng2021simple}
D.~Feng, Y.~Zhou, C.~Xu, M.~Tomizuka \emph{et~al.}, ``{A simple and efficient multi-task network for 3d object detection and road understanding},'' in \emph{IEEE/RSJ International Conference on Intelligent Robots and Systems (IROS)}.\hskip 1em plus 0.5em minus 0.4em\relax IEEE, 2021, pp. 7067--7074.

\bibitem{chen2024joint}
X.~Chen, J.~Cui, Y.~Liu, X.~Zhang, J.~Sun, R.~Ai, W.~Gu, J.~Xu, and H.~Lu, ``Joint scene flow estimation and moving object segmentation on rotational lidar data,'' \emph{IEEE Transactions on Intelligent Transportation Systems}, 2024.

\bibitem{elamir2022deep}
M.~S. Elamir, H.~Gotzig, R.~Z{\"o}llner, and P.~M{\"a}der, ``A deep learning approach for direction of arrival estimation using automotive-grade ultrasonic sensors,'' in \emph{Journal of Physics: Conference Series}, vol. 2234, no.~1.\hskip 1em plus 0.5em minus 0.4em\relax IOP Publishing, 2022, p. 012009.

\bibitem{kumar2021omnidet}
V.~R. Kumar, S.~Yogamani, H.~Rashed, G.~Sitsu \emph{et~al.}, ``{Omnidet: Surround view cameras based multi-task visual perception network for autonomous driving},'' \emph{IEEE Robotics and Automation Letters}, vol.~6, no.~2, pp. 2830--2837, 2021.

\bibitem{liu2022convnet}
Z.~Liu, H.~Mao, C.-Y. Wu, C.~Feichtenhofer \emph{et~al.}, ``{A convnet for the 2020s},'' in \emph{Proceedings of the IEEE/CVF conference on computer vision and pattern recognition}, 2022, pp. 11\,976--11\,986.

\bibitem{paszke2017automatic}
A.~Paszke, S.~Gross, S.~Chintala, G.~Chanan \emph{et~al.}, ``{Automatic differentiation in pytorch},'' 2017.

\bibitem{kingma2014adam}
D.~P. Kingma and J.~Ba, ``{Adam: A method for stochastic optimization},'' \emph{arXiv preprint arXiv:1412.6980}, 2014.

\bibitem{loshchilov2016sgdr}
I.~Loshchilov and F.~Hutter, ``{Sgdr: Stochastic gradient descent with warm restarts},'' \emph{arXiv preprint arXiv:1608.03983}, 2016.

\bibitem{lin2017focal}
T.-Y. Lin, P.~Goyal, R.~Girshick, K.~He \emph{et~al.}, ``{Focal loss for dense object detection},'' in \emph{Proceedings of the IEEE international conference on computer vision}, 2017, pp. 2980--2988.

\bibitem{girshick2015fast}
R.~Girshick, ``{Fast r-cnn},'' in \emph{Proceedings of the IEEE international conference on computer vision}, 2015, pp. 1440--1448.

\bibitem{wu2023transformation}
H.~Wu, C.~Wen, W.~Li, X.~Li, R.~Yang, and C.~Wang, ``Transformation-equivariant 3d object detection for autonomous driving,'' in \emph{Proceedings of the AAAI Conference on Artificial Intelligence}, vol.~37, no.~3, 2023, pp. 2795--2802.

\bibitem{shi2019part}
S.~Shi, Z.~Wang, X.~Wang, and H.~Li, ``{Part-aˆ 2 net: 3d part-aware and aggregation neural network for object detection from point cloud},'' \emph{arXiv preprint arXiv:1907.03670}, vol.~2, no.~3, 2019.

\bibitem{zhou2020cylinder3d}
H.~Zhou, X.~Zhu, X.~Song, Y.~Ma \emph{et~al.}, ``{Cylinder3d: An effective 3d framework for driving-scene lidar semantic segmentation},'' \emph{arXiv preprint arXiv:2008.01550}, 2020.

\bibitem{cortinhal2020salsanext}
T.~Cortinhal, G.~Tzelepis, and E.~Erdal~Aksoy, ``{Salsanext: Fast, uncertainty-aware semantic segmentation of lidar point clouds},'' in \emph{Advances in Visual Computing: 15th International Symposium, ISVC, Proceedings, Part II 15}.\hskip 1em plus 0.5em minus 0.4em\relax Springer, 2020, pp. 207--222.

\bibitem{kim2022rvmos}
J.~Kim, J.~Woo, and S.~Im, ``{RVMOS: Range-View Moving Object Segmentation Leveraged by Semantic and Motion Features},'' \emph{IEEE Robotics and Automation Letters}, vol.~7, no.~3, pp. 8044--8051, 2022.

\bibitem{mersch2022receding}
B.~Mersch, X.~Chen, I.~Vizzo, L.~Nunes \emph{et~al.}, ``{Receding moving object segmentation in 3d lidar data using sparse 4d convolutions},'' \emph{IEEE Robotics and Automation Letters}, vol.~7, no.~3, pp. 7503--7510, 2022.

\bibitem{nni2021}
\BIBentryALTinterwordspacing
{Microsoft}, ``{Neural Network Intelligence},'' 1 2021. [Online]. Available: \url{https://github.com/microsoft/nni}
\BIBentrySTDinterwordspacing

\end{thebibliography}

\begin{IEEEbiography}[{\includegraphics[width=1in,height=1.25in,clip,keepaspectratio]{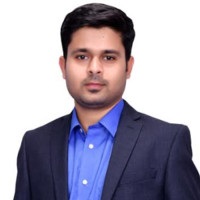}}]{Sambit Mohapatra} 
is currently working as a software engineer at Valeo, Germany and also pursuing a PhD in efficient neural architectures and neuromorphic computing for autonomous driving systems at the Technische Universität Ilmenau, Germany. His research includes application of efficent architectures, graph neural networks and spiking neural networks for real time perception using ranging sensors on embedded systems and neuromoprhic hardware. He is an author of 4 publications and 4 filed patents. He also serves as a reviewer for several IEEE automotive conferences including ITSC and IV.
\end{IEEEbiography}

\begin{IEEEbiography}[{\includegraphics[width=1in,height=1.25in,clip,keepaspectratio]{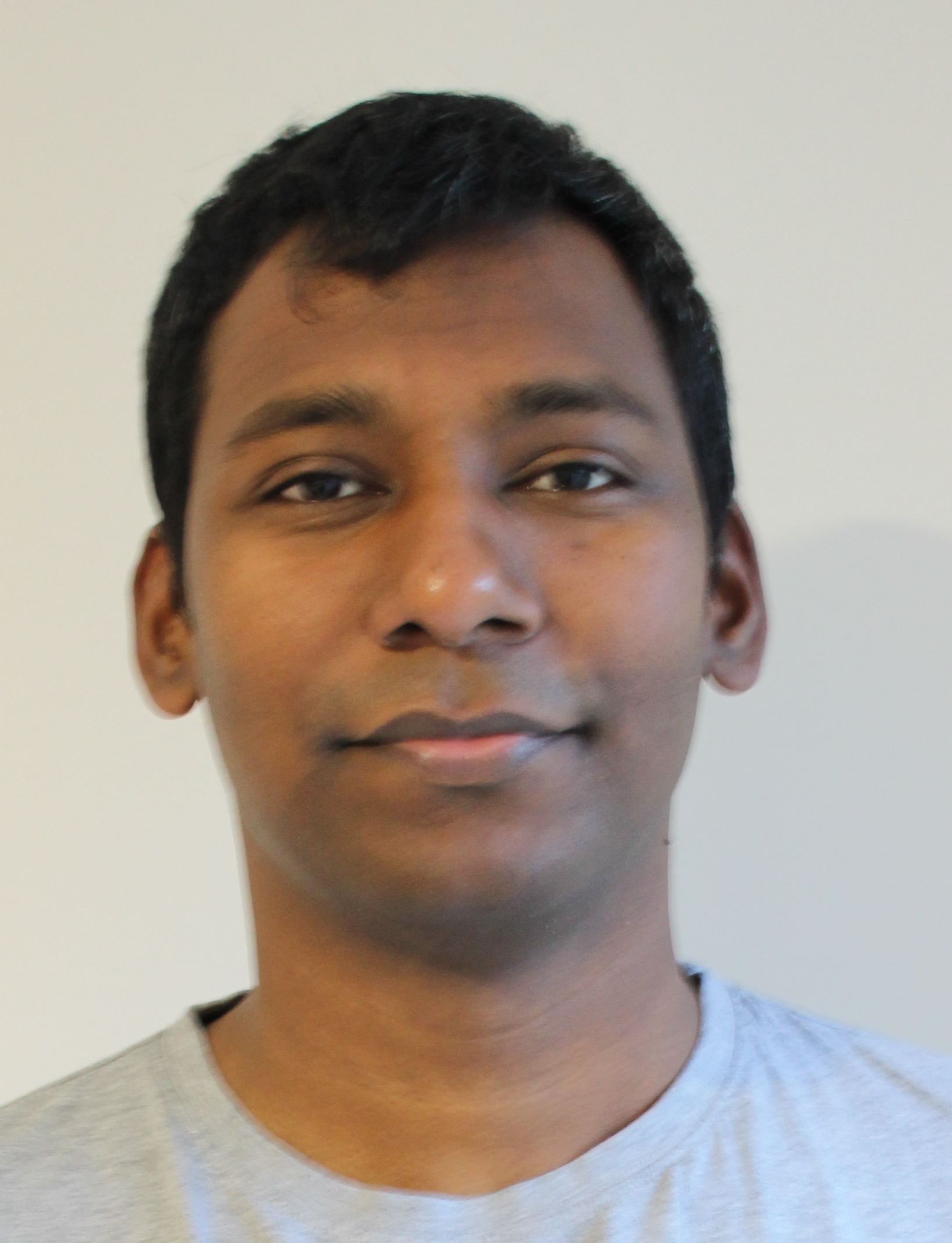}}]{Senthil Yogamani} holds an engineering director position at Qualcomm and leads the data-centric AI department for autonomous driving. He has over 19 years of experience in computer vision and machine learning including 16 years of experience in industrial automotive systems. He is an author of 150+ publications which have 8400 citations and an inventor of 200+ filed patents. He serves on the editorial board of various leading IEEE automotive conferences including ITSC and IV.
\end{IEEEbiography}

\begin{IEEEbiography}[{\includegraphics[width=1in,height=1.25in,trim={ 0 1cm 0 0},clip,keepaspectratio]{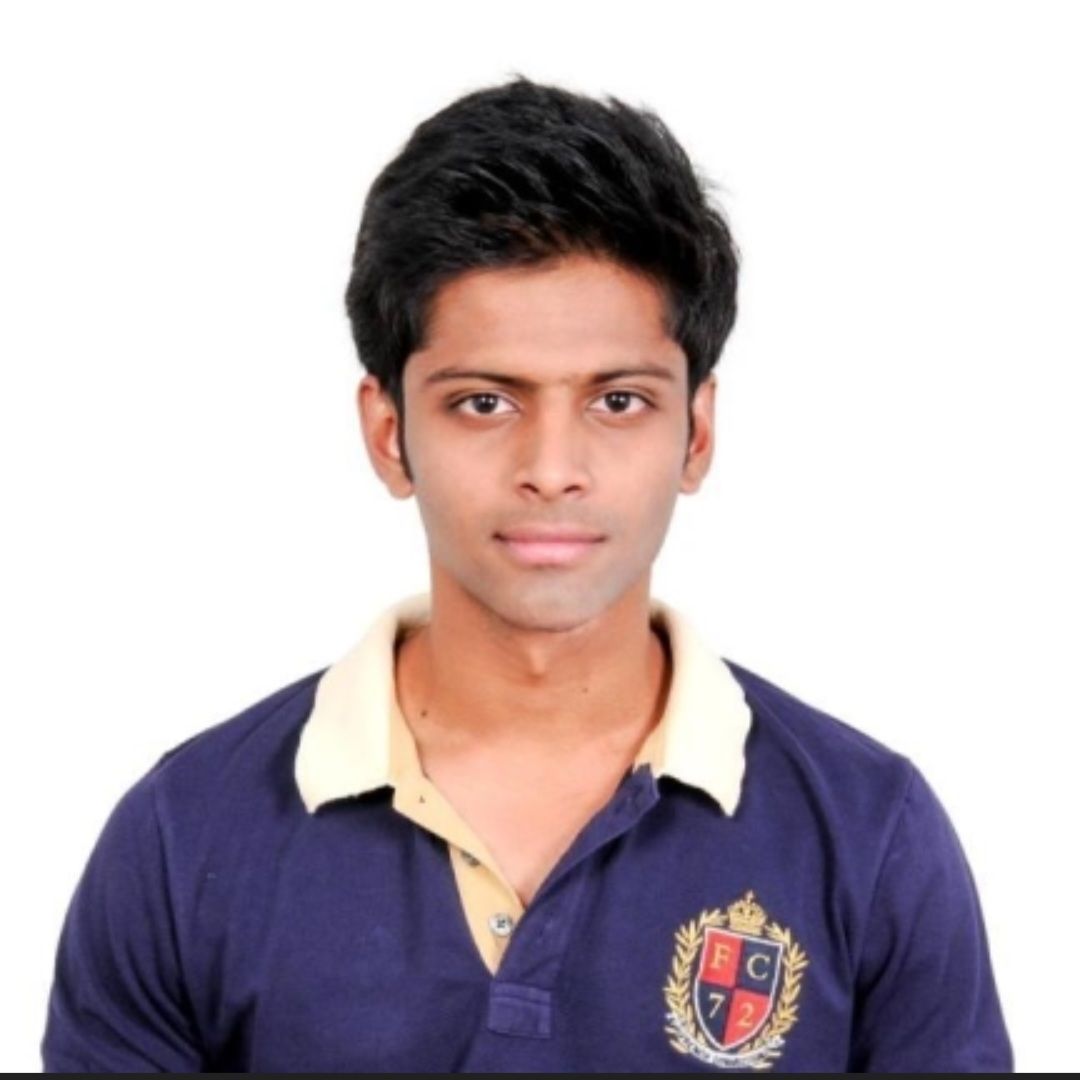}}]{Varun Ravi Kumar} holds a Master’s degree in Embedded Systems from Technische Universität Chemnitz, Germany and a Ph.D. in Artificial Intelligence (AI) from Technische Universität Ilmenau, Germany. He currently holds a position at Qualcomm, where he serves as a Staff Research Scientist/Manager. In this role, he leads Deep Learning (DL) teams focused on Autonomous Driving. For over 8+ years, his research has concentrated on developing DL based self-supervised algorithms using neural networks. He is the author of 32 scientific articles, published in prestigious journals and conferences such as CVPR, ICCV, ICRA, IROS, and WACV, with 1109 citations, an h-index of 19, and an i10-index of 25. Additionally, he is a prolific inventor of more than 100+ filed patents.
\end{IEEEbiography}
\vspace{-6cm}

\begin{IEEEbiography}[{\includegraphics[width=1in,height=1.25in,clip,keepaspectratio]{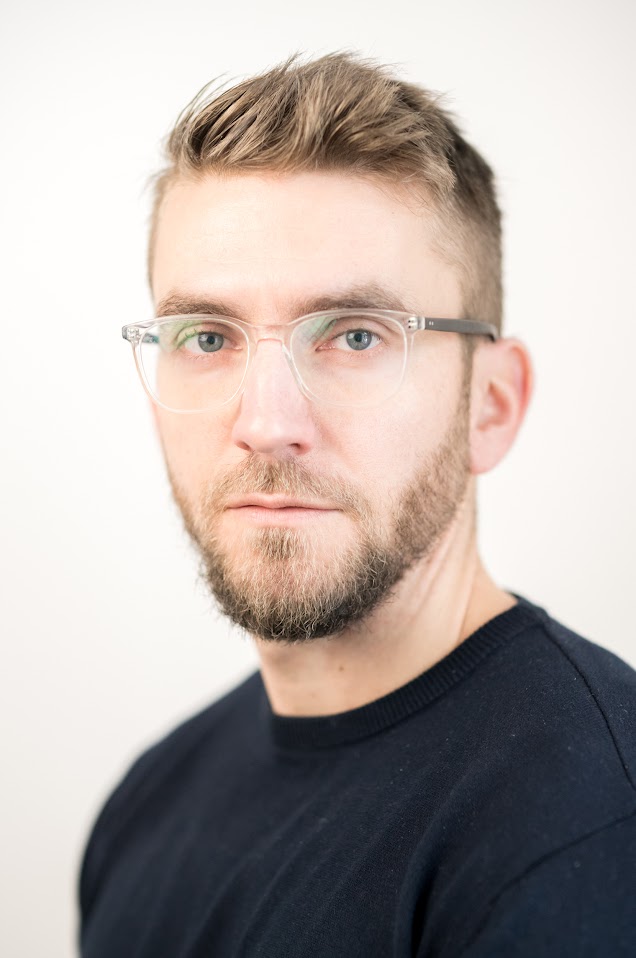}}]{Stefan Milz} received  his  Ph.D.  degree in  Physics from the Technical University of Munich. He is Managing Director of Spleenlab.ai,  a  self-founded  machine learning  company focusing on safety-critical mobility applications (UAV, Automated Driving, Air-Taxis) deploying SLAM, sensor-fusion, perception functions into the real world with regard to saftey standards. He is also a research fellow at  the  TU-Ilmenau. He is  the  author  and co-author of more than 60 patents and more than 60 publications.
\end{IEEEbiography}
\vspace{-6cm}

\begin{IEEEbiography}[{\includegraphics[width=1in,height=1.15in,clip,keepaspectratio]{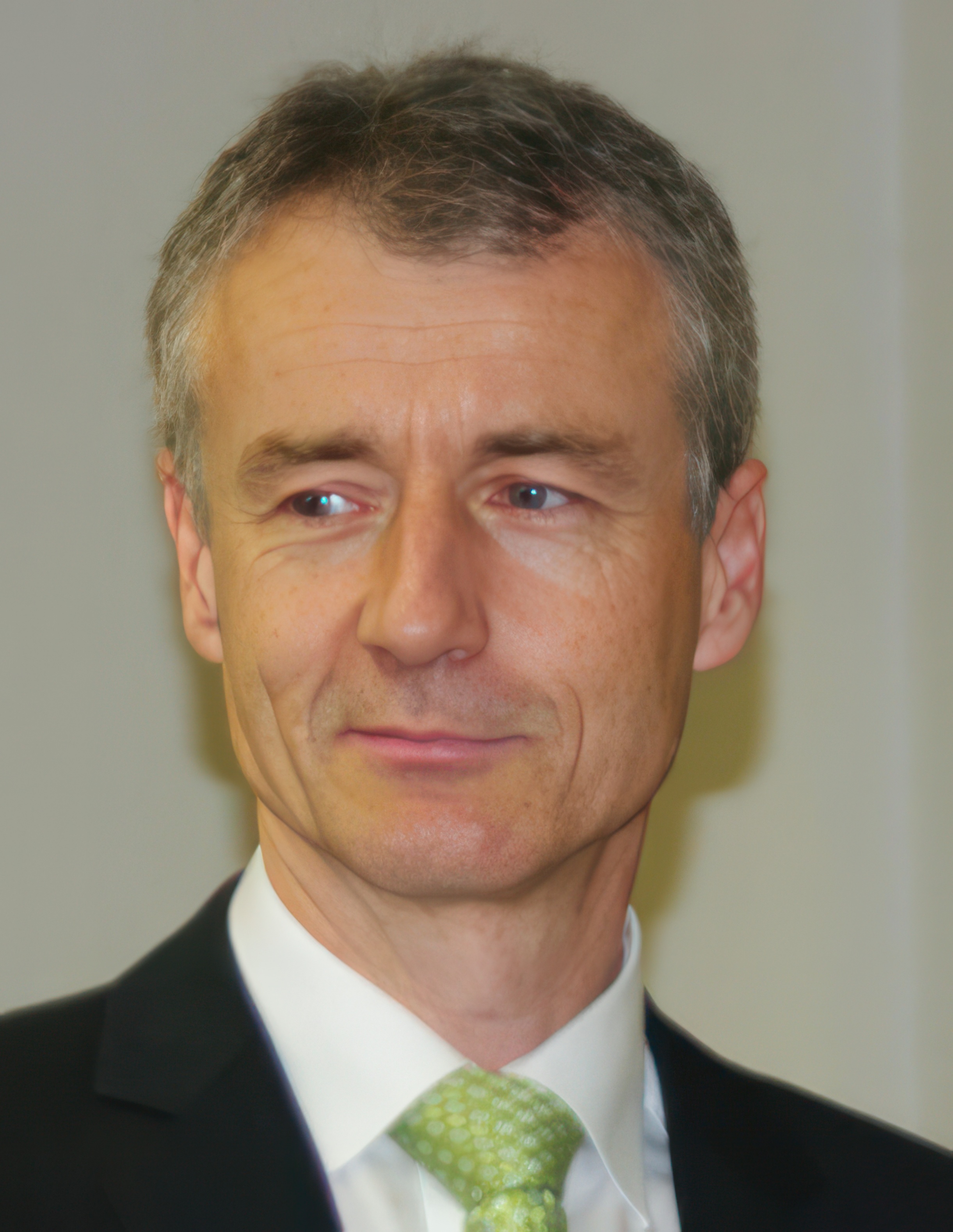}}]{Heinrich Gotzig} works as a director for advanced development and expertise at Valeo, Germany. With a vast experience of over 2 decades, he leads a large pool of researchers at Valeo in the area of sensor technologies and efficient neural architectures for advanced driving assistance systems. His research interests include sensor modeling, driving assistance algorithms and new sensor research.
\end{IEEEbiography}
\vspace{-6cm}

\begin{IEEEbiography}[{\includegraphics[width=1in,height=1.25in,clip,keepaspectratio]{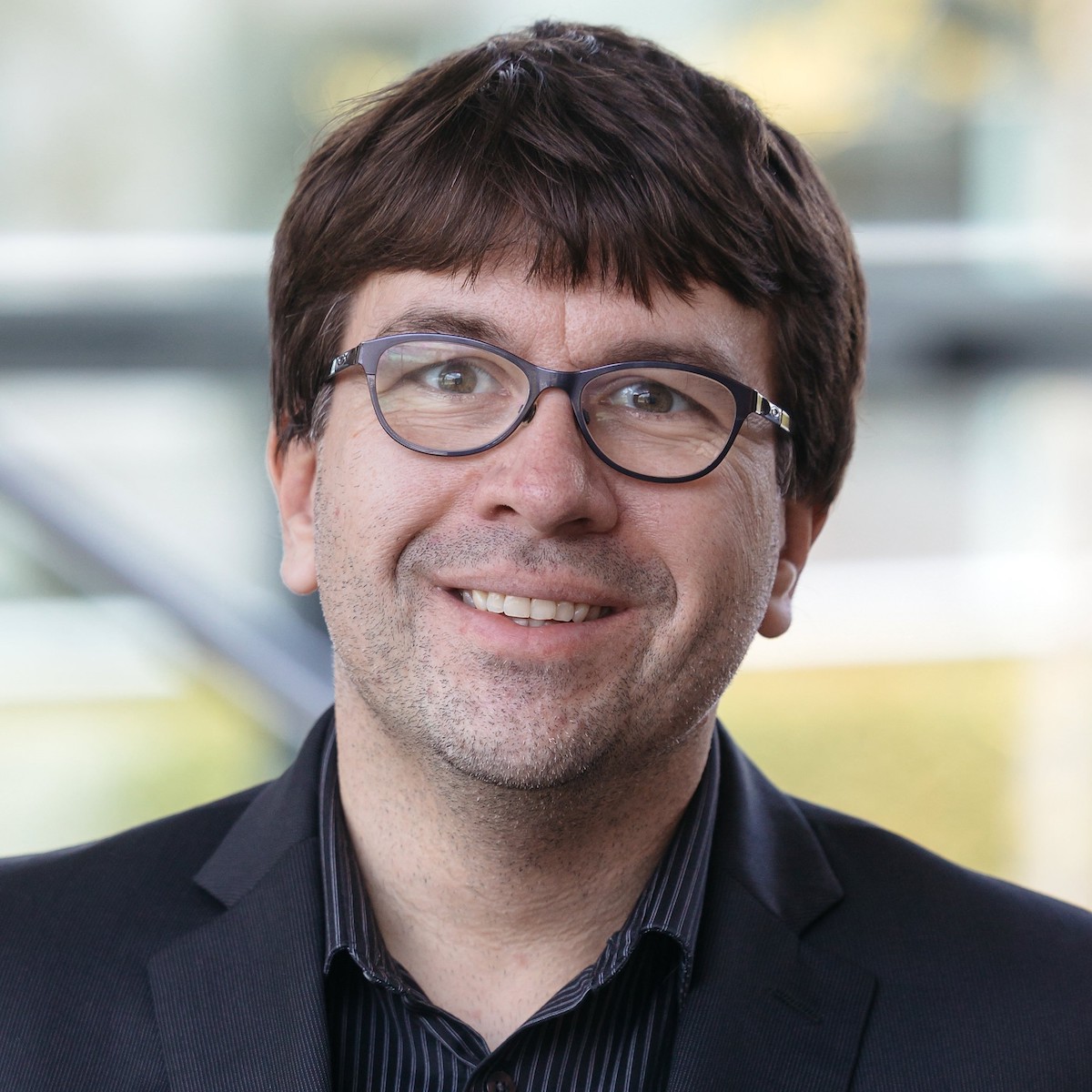}}]{Patrick Maeder} is a Professor at the Technische Universität Ilmenau,
Germany for Data-intensive Systems and Visualization. His research
inter- ests include machine learning, software engineering, and safety
engineering. He worked as a consultant for the EXTESSY AG, Wolfsburg, as
postdoctoral fellow with the Institute for Systems Engineering
and Automation (SEA) of the Johannes Kepler University Linz, Austria and
as postdoctoral researcher with the Software and Requirements
Engineering Center, DePaul University Chicago, USA.
\end{IEEEbiography}

\end{document}